%% file: acl_latex.tex
\documentclass[11pt]{article}
\usepackage[final]{acl}
\usepackage{times}
\usepackage{latexsym}
\usepackage[T1]{fontenc}
\usepackage[utf8]{inputenc}
\usepackage{microtype}
\usepackage{inconsolata}
\usepackage{amsmath,amssymb,amsthm,mathtools}
\usepackage{booktabs}
\usepackage{multirow}
\usepackage{makecell}
\usepackage{graphicx}
\usepackage{subcaption}
\usepackage[table]{xcolor}
\usepackage{enumitem}
\usepackage{array}
\usepackage{tabularx}
\usepackage{url}
\usepackage{placeins}
\usepackage{algorithm}
\usepackage{algpseudocode}
\usepackage{hyperref}
\newcommand{\method}{FCCA}
\newcommand{\R}{\mathbb{R}}
\newcommand{\bbE}{\mathbb{E}}
\newcommand{\rank}{\operatorname{rank}}
\newcommand{\Tr}{\operatorname{Tr}}
\newcommand{\diag}{\operatorname{diag}}
\newcommand{\tgrad}{\widetilde{G}}
\newcommand{\core}{R}
\newcommand{\takeaway}[1]{\textbf{\emph{#1}}}
\newtheorem{theorem}{Theorem}

\title{Frozen Cores Need Task Signal:\\Fisher-Whitened Cross-Covariance for Parameter-Efficient Adaptation}
\author{Wentao Ye\textsuperscript{1,2,*}, Zhanming Shen\textsuperscript{1,*}, 
   Zhiqing Xiao\textsuperscript{1}, Yao Ding\textsuperscript{2}, Haobo Wang\textsuperscript{1,\ensuremath{\dagger}},
   Gang Chen\textsuperscript{1}\\
   \textsuperscript{1} Zhejiang University \textsuperscript{2}Hunyuan, Tencent\\
   \texttt{\{yewt01, z.shen, wanghaobo\}@zju.edu.cn}
   }

\begin{document}
\maketitle

\begingroup
\renewcommand{\thefootnote}{\fnsymbol{footnote}}
\footnotetext[1]{Equal contribution.}
\footnotetext[2]{Corresponding author.}
\endgroup

\begin{abstract}
Parameter-efficient fine-tuning is usually framed as a question of \emph{how many} parameters to update. Under a severe trainable-state budget, however, \emph{where} those coefficients act is equally consequential. We study this choice through \emph{frozen-core adaptation}: a calibration pass fixes left and right bases for each weight matrix, and fine-tuning optimizes only an $r\times r$ core. This removes the ability of trainable factors to repair a poor initial span and makes subspace quality directly observable. We introduce \method{}, which estimates the signed input--error cross-covariance, whitens it with diagonal Fisher moments, truncates it in the resulting local metric, maps the selected directions back, and applies thin QR to obtain stable core coordinates. Under a matched $r^2$ budget, we compare eight basis constructors on 11 tasks, four model settings, and three seeds. On Qwen2.5-3B, \method{} reaches an 83.0 macro-average, 2.3 points above the next-best matched-budget constructor, and exceeds its unwhitened RawGrad control on all 11 tasks. It ranks first at all three Qwen scales and finishes within 0.13 points of the best method on Llama-3.2-1B. Controlled ablations show gains of 2.7--17.2 points from whitening and identify QR as necessary for stable core optimization in the tested regime. Finally, \method{} comes within 0.32 and 0.23 average points of LoRA and DoRA while optimizing 36.9K rather than roughly 7.4M parameters. These results show that a carefully selected fixed span can recover most of the benefit of movable low-rank factors at a much smaller trainable and optimizer-state cost.\footnote{The source code and data can be found \href{https://github.com/yyy01/FCCA}{here}.}
\end{abstract}

\begingroup
\setlength{\parskip}{0pt}
\section{Introduction}
Parameter-efficient fine-tuning (PEFT) is commonly presented as a parameter-count problem: keep the pretrained model frozen, introduce a small adapter, and optimize as few new values as possible. This view explains much of LoRA's appeal \citep{hu2022lora}, but it leaves a second question implicit. When the budget is extremely small, the location of the trainable coefficients can matter as much as their number. A handful of coefficients placed in a task-relevant update span may be useful; the same coefficients placed in directions inherited from pretraining, random initialization, or poorly scaled gradients may be nearly inert.

Low-rank adapters normally hide this distinction. In LoRA, two trainable factors jointly determine the row and column spaces of the update and the coefficients within those spaces. They therefore perform two jobs at once: they must discover a useful subspace and optimize the update inside it. End-task accuracy conflates these effects. A favorable initialization may help, but an initially weak span can also rotate during fine-tuning, making it difficult to tell whether a method selected good directions or merely recovered from bad ones. This ambiguity is especially important when data, steps, optimizer state, or communication are constrained, because the opportunity to revise a span is limited.

\begin{figure*}[htb]
\centering
\begin{subfigure}[t]{0.27\textwidth}
  \centering
  \includegraphics[width=\linewidth]{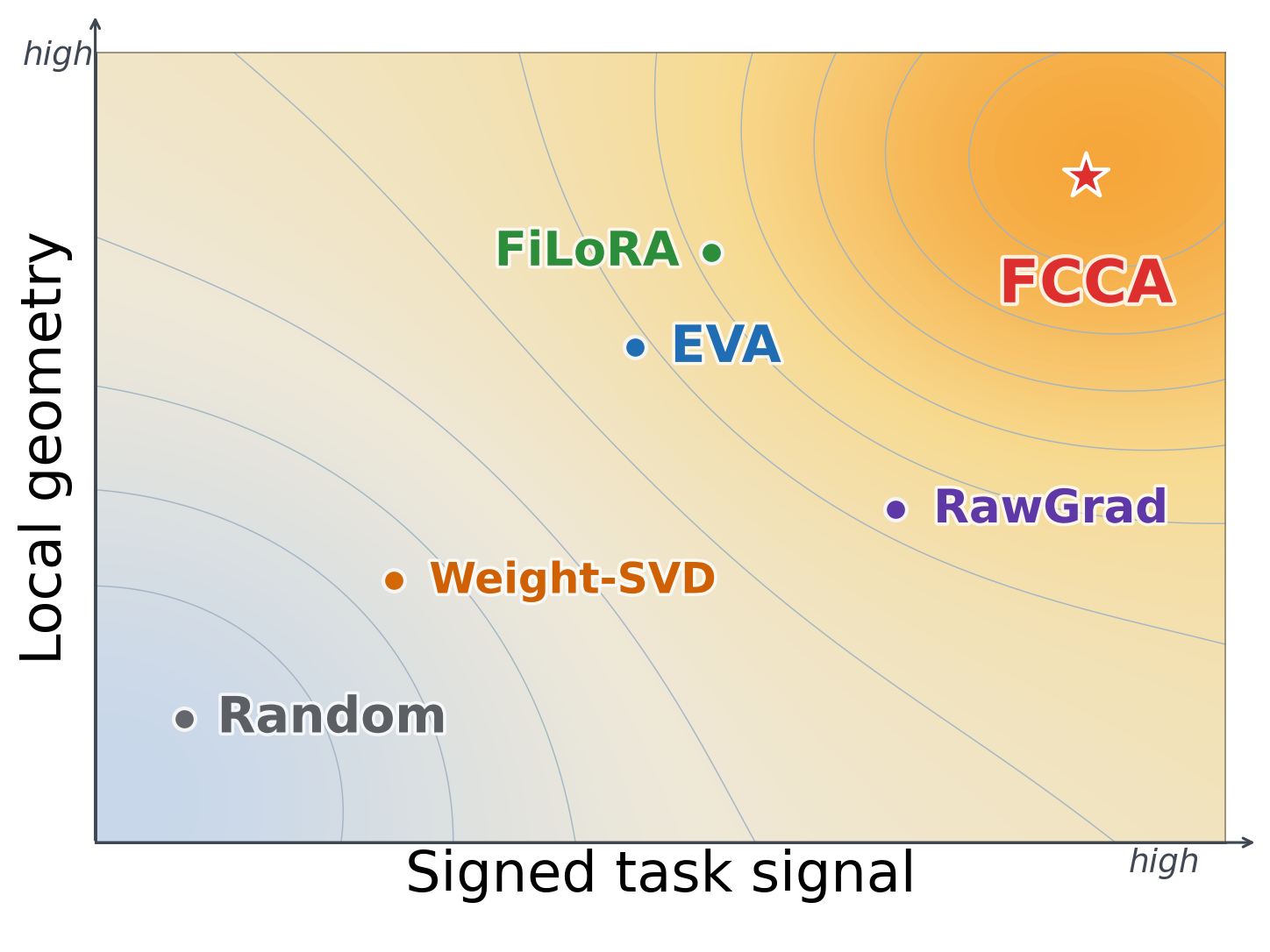}
  \caption{Conceptual design space.}
  \label{fig:method-design}
\end{subfigure}\hfill
\begin{subfigure}[t]{0.7\textwidth}
  \centering
  \includegraphics[width=\linewidth]{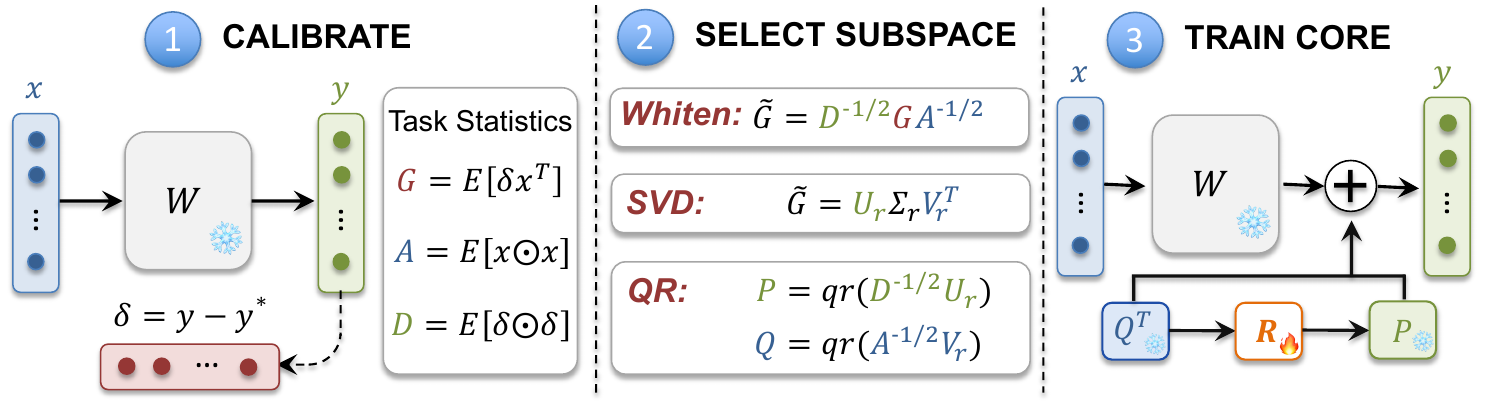}
  \caption{FCCA pipeline.}
  \label{fig:method-pipeline}
\end{subfigure}
\caption{Panel (\subref{fig:method-design}) sketches the design space of frozen-span construction: useful spans should preserve both signed task signal and local geometry, and the labeled positions indicate which information each constructor retains before truncation. Panel (\subref{fig:method-pipeline}) shows FCCA: calibration estimates the signed cross-covariance and diagonal Fisher moments; whitening imposes the rank constraint in Fisher coordinates; inverse scaling and thin QR map the retained directions back to stable frozen bases; only the core $R$ is trained.}
\label{fig:method}
\end{figure*}

We isolate subspace selection with \emph{frozen-core adaptation}. For a pretrained matrix $W\in\R^{m\times n}$, calibration fixes bases $P\in\R^{m\times r}$ and $Q\in\R^{n\times r}$, and fine-tuning optimizes only the core in $PRQ^\top$. Matching the model, target matrices, rank, and $r^2$ trainable budget makes basis choice directly comparable. As a diagnostic, the protocol reveals how much downstream value is already present before a span can rotate; as an adaptation model, it represents settings where gradients, optimizer states, or transmitted updates must remain tiny. The fixed bases still have a storage cost, so the efficiency claim concerns trainable and optimizer state rather than a universally tiny task artifact.

What information should define such a span? Weight-spectrum bases emphasize directions prominent in pretraining but never observe the downstream loss \citep{balazy2024loraxs,wang2024milora}. Activation- and marginal-Fisher methods observe the calibration distribution, yet they do not retain which activation produced which signed error \citep{paischer2025eva,han2026filora}. A raw task-gradient basis preserves that pairing, but an ordinary SVD imposes rank in Euclidean coordinates: high-variance input or error dimensions can dominate even when movement along them is locally expensive. These constructors therefore capture different pieces of the same object. The missing link is a way to preserve the signed task direction while judging it in the local geometry of the loss, rather than yet another source of marginal energy. A strong frozen span should do both before the rank constraint discards information.

Figure~\ref{fig:method} organizes this design space and summarizes \method{} (Fisher-whitened cross-covariance adaptation). For layer inputs $x$ and backpropagated output errors $\delta$, calibration estimates the signed cross-moment $G=\bbE[\delta x^\top]$ together with input- and output-side second moments $A$ and $D$. We first form $\widetilde G=D^{-1/2}GA^{-1/2}$ and apply the rank constraint in these Fisher-scaled coordinates. Its leading singular directions solve a local rank-constrained objective; mapping them back gives raw bases in the original parameter coordinates. Thin QR then orthonormalizes those bases without changing the representable update family. This order is essential: whitening after truncation could rescale a chosen span, but it could not recover directions already discarded by the unscaled SVD. QR addresses a separate problem by presenting the retained span in coordinates that a common core optimizer can train.

Across 11 tasks and four model settings, \method{} ranks first on Qwen2.5-1.5B, 3B, and 7B and is essentially tied for first on Llama-3.2-1B. At 3B, it reaches an 83.0 mean, leads the next-best matched-budget constructor by 2.3 points, and exceeds RawGrad on every task. Whitening helps all six mechanism tasks; calibration-size experiments show that useful directions emerge from modest task-matched data; and rank, module, QR, and dense-curvature controls separate span selection from coordinate conditioning. The fixed span also remains within 0.32 points of LoRA and 0.23 points of DoRA while optimizing roughly $200\times$ fewer values.

This work makes four contributions. (1) We formulate frozen-subspace selection in a local Fisher metric and derive the truncated whitened cross-covariance as its rank-$r$ solution. (2) We introduce a unified frozen-core evaluation across eight constructors, 11 tasks, three Qwen scales, a Llama architecture, and three seeds, exposing basis quality under an exactly matched $r^2$ trainable budget. (3) Controlled interventions identify the roles of signed task signal, whitening, calibration coverage, span size, and adapted modules. (4) Additional QR and retuned no-QR controls show that even a useful span must still be expressed in coordinates the optimizer can train. Together, these results sharpen the main conclusion: low-rank adaptation benefits from choosing directions in the metric where information will later be discarded.
\endgroup
\vspace{-0.25em}
\section{Related Work}
\paragraph{Parameter-efficient adaptation.}
Adapters, prompt and prefix tuning, BitFit, and (IA)$^3$ reduce trainable state through different parameterizations \citep{houlsby2019parameter,li2021prefixtuning,lester2021power,benzaken2022bitfit,liu2022ia3}. LoRA trains both low-rank factors \citep{hu2022lora}; DoRA separates direction from magnitude \citep{liu2024dora}; PiSSA initializes trainable factors from principal weight components \citep{meng2024pissa}; and AdaLoRA allocates rank dynamically \citep{zhang2023adalora}. In contrast, our controlled setting freezes both bases and trains only a square core, isolating the quality of the selected subspace.

\paragraph{Fixed or data-derived subspaces.}
VeRA uses shared random low-rank matrices with lightweight trainable scaling \citep{kopiczko2024vera}. LoRA-XS places a small trainable matrix between weight-SVD bases \citep{balazy2024loraxs}, whereas MiLoRA emphasizes minor singular components \citep{wang2024milora}. EVA derives bases from activation variance \citep{paischer2025eva}; FiLoRA uses Fisher-guided marginal subspaces \citep{han2026filora}; LoRA-GA and gradient-compression methods exploit task gradients in trainable-factor settings \citep{wang2024loraga,hao2024flora}. We instantiate each applicable idea in the same frozen-basis, trainable-core form so that performance reflects subspace choice instead of differences in native parameterization.

\paragraph{Curvature and natural gradients.}
Natural-gradient and K-FAC methods rescale optimization by local parameter geometry \citep{amari1998natural,martens2015optimizing}. \method{} applies this principle to \emph{subspace selection}: it preserves the signed input--error cross-covariance and normalizes both sides before the rank constraint is imposed. This is distinct from selecting input and output eigenspaces independently.

\begin{table}[htb]
\centering
\scriptsize
\setlength{\tabcolsep}{3.2pt}
\begin{tabular}{lccc}
\toprule
Constructor & Task signal & Geometry & Source \\
\midrule
Random & no & no & random draw \\
LoRA-XS/MiLoRA & no & weight & $W$ spectrum \\
EVA-core & indirect & input & activations \\
FiLoRA-core & indirect & input+output & $A,D$ \\
RawGrad & yes & no & $G$ \\
\textbf{FCCA} & yes & input+output & $G,A,D$ \\
\bottomrule
\end{tabular}
\caption{Information used by the matched-budget basis constructors. ``Indirect'' means the data distribution is observed without the signed input--error cross term.}
\label{tab:basis-signals}
\end{table}

Table~\ref{tab:basis-signals} summarizes the information available to each constructor. Weight-spectrum methods retain directions prominent in pretraining but never observe the downstream task. EVA observes task inputs, and FiLoRA observes input and output marginals, but neither retains the signed pairing between an activation and the error it produced. RawGrad retains that pairing in a Euclidean metric. Among the tested methods, FCCA alone combines the signed cross term with both marginal geometries before truncation.

\section{Method}
\subsection{Frozen-core adaptation}
For a frozen linear map $y=Wx$, we use $y=(W+sP\core Q^\top)x$. The columns of $P\in\R^{m\times r}$ and $Q\in\R^{n\times r}$ are orthonormal, $s$ is a fixed adapter scale, and only $\core\in\R^{r\times r}$ is trainable. Calibration fixes the bases, and training begins from $\core=0$. Across $L$ target matrices, the trainable budget is exactly $Lr^2$, independent of matrix width, and the update family $\mathcal S(P,Q)=\{P\core Q^\top:\core\in\R^{r\times r}\}$ cannot rotate during optimization.

\paragraph{A directly observable criterion for subspace quality.}
Let $G=\nabla_W\mathcal L$ be the task gradient of the frozen matrix. At the zero core, $\nabla_R\mathcal L=sP^\top GQ$, so a gradient step of size $\gamma$ reduces the linearized loss in proportion to $\gamma s^2\|P^\top GQ\|_F^2$. Because $s$ is shared across constructors, the projected norm $\|P^\top GQ\|_F$ is a directly observable criterion for basis quality. A useful frozen subspace must capture signed task gradient in both its left and right spans. High activation variance, large weight singular values, or large marginal Fisher eigenvalues alone do not ensure this. Unlike trainable LoRA factors, a frozen basis cannot rotate toward a missed direction during fine-tuning.

The Euclidean criterion is nevertheless coordinate dependent: high-variance activation or error coordinates can dominate $G$ even when movement along them is locally expensive. We therefore impose the rank constraint in a Fisher-scaled metric rather than truncating the raw gradient directly.

For each calibration token, let $x\in\R^n$ be the matrix input and $\delta=\partial\mathcal L/\partial y\in\R^m$ the backpropagated output error. We estimate the signed cross moment $G=\bbE[\delta x^\top]$ and diagonal moments $a=\bbE[x^{\odot2}]$ and $d=\bbE[\delta^{\odot2}]$. The default factors are $A=\diag(a)+\epsilon I$ and $D=\diag(d)+\epsilon I$. This diagonal estimator avoids fitting two dense width-by-width covariances from only a few hundred calibration examples.

\subsection{Fisher-whitened cross-covariance}
Consider the local rank-constrained objective
\begin{equation}
\label{eq:local}
\begin{aligned}
\min_{\rank(\Delta W)\le r}\quad
 &\langle G,\Delta W\rangle \\
 &+\frac{1}{2\eta}
   \left\|D^{1/2}\Delta W A^{1/2}\right\|_F^2.
\end{aligned}
\end{equation}
The quadratic term defines a Kronecker-factored Fisher metric for the linear layer. Defining $\tgrad=D^{-1/2}GA^{-1/2}$ moves the signed task gradient into Fisher-whitened coordinates \emph{before} any low-rank truncation.

\begin{theorem}
Let $A$ and $D$ be positive definite, and let $[\tgrad]_r$ denote a best rank-$r$ approximation to $\tgrad$. A minimizer of Eq.~\eqref{eq:local} is
\begin{equation}
 \Delta W^*=-\eta D^{-1/2}[\tgrad]_rA^{-1/2}.
 \label{eq:solution}
\end{equation}
\end{theorem}

The proof changes variables to Fisher-whitened coordinates and applies the Eckart--Young--Mirsky theorem \citep{eckart1936approximation,mirsky1960symmetric}; Appendix~\ref{app:proof} gives the derivation. If $\tgrad=U\Sigma V^\top$, Eq.~\eqref{eq:solution} yields raw bases $P_0=D^{-1/2}U_r$ and $Q_0=A^{-1/2}V_r$. RawGrad is the exact no-whitening control obtained by factorizing $G$ instead. Marginal-curvature methods select directions from $A$ and $D$ separately and therefore lose the signed activation--error pairing.

\subsection{QR as a coordinate reparameterization}
Inverse whitening can leave $P_0$ and $Q_0$ severely anisotropic. We compute thin QR factorizations $P_0=PT_P$ and $Q_0=QT_Q$ and train the orthonormal-basis adapter $P\core Q^\top$. Because $P_0ZQ_0^\top=P(T_PZT_Q^\top)Q^\top$, QR preserves the representable update family and changes only the coordinates of the square core. The controlled ablation in Table~\ref{tab:design-summary} shows that these coordinates are essential under the shared learning rate.

\begin{algorithm}[htb]
\small
\caption{FCCA basis construction and core training}
\label{alg:fcca}
\begin{algorithmic}[1]
\Require frozen model, calibration set $\mathcal D_c$, target matrices $\mathcal M$, rank $r$
\For{$\ell\in\mathcal M$}
  \State initialize $G_\ell,a_\ell,d_\ell\leftarrow 0$
\EndFor
\For{calibration batch $B\subset\mathcal D_c$}
  \State run one forward/backward pass
  \For{$\ell\in\mathcal M$}
    \State $G_\ell\mathrel{+}=\sum_t\delta_{\ell,t}x_{\ell,t}^\top$
    \State $a_\ell\mathrel{+}=\sum_t x_{\ell,t}^{\odot2}$; $d_\ell\mathrel{+}=\sum_t\delta_{\ell,t}^{\odot2}$
  \EndFor
\EndFor
\For{$\ell\in\mathcal M$}
  \State $A_\ell\leftarrow\diag(a_\ell)+\epsilon I$; $D_\ell\leftarrow\diag(d_\ell)+\epsilon I$
  \State $\tgrad_\ell\leftarrow D_\ell^{-1/2}G_\ell A_\ell^{-1/2}$
  \State $U_r,\Sigma_r,V_r\leftarrow\operatorname{SVD}_r(\tgrad_\ell)$
  \State $P_0\leftarrow D_\ell^{-1/2}U_r$; $Q_0\leftarrow A_\ell^{-1/2}V_r$
  \State thin QR of $P_0,Q_0$; attach a zero core $R_\ell$
\EndFor
\State discard calibration statistics and train only $\{R_\ell\}$
\end{algorithmic}
\end{algorithm}

\subsection{Construction, training, and storage costs}
Algorithm~\ref{alg:fcca} summarizes construction and training. For each $m\times n$ target matrix, calibration accumulates one $m\times n$ cross moment and $m+n$ diagonal moments, followed by a truncated rank-$r$ SVD. Our implementation hooks all target layers in one grouped forward/backward pass and stores the statistics on CPU. Cross moments can dominate host memory, and the backward pass is more expensive than forward-only activation collection. This construction cost is paid once per task. Streamed or randomized accumulation could reduce the footprint but is not evaluated here.

After construction, gradients and optimizer states are required for only $r^2$ values per target matrix, versus $r(m+n)$ values for a trainable-basis rank-$r$ adapter. The fixed bases still occupy $r(m+n)$ storage when retained separately for each task. Our efficiency claim therefore concerns trainable and optimizer state: an unmerged adapter must store $P$ and $Q$, whereas a merged update requires a task-specific copy of the base model.

\section{Experimental Setup}
\paragraph{Models and tasks.}
Qwen2.5-3B-Instruct is the primary model; Qwen2.5-1.5B/7B-Instruct and Llama-3.2-1B-Instruct provide scale and architecture checks \citep{yang2024qwen25,grattafiori2024llama3}. The 11-task suite contains SVAMP and GSM8K \citep{patel2021svamp,cobbe2021gsm8k}; SST-2, QNLI, CoLA, RTE, and MRPC \citep{wang2018glue}; and ARC-Challenge, OpenBookQA, HellaSwag, and WinoGrande \citep{clark2018arc,mihaylov2018openbookqa,zellers2019hellaswag,sakaguchi2020winogrande}. We evaluate fixed subsets of 300 examples per math task and 500 per remaining task. Appendix~\ref{app:protocol} and Table~\ref{tab:datasets} give dataset configurations, sizes, and training lengths.

\paragraph{Common adapter protocol.}
Every matched-budget method uses rank $r=16$, adapts all $q/k/v/o$ projections, runs in bf16 with effective batch size 8, initializes the core to zero, and trains one $r\times r$ core per target matrix. Qwen2.5-3B therefore has 36,864 trainable parameters. We compare \method{}, RawGrad, FiLoRA-core, MiLoRA-core, LoRA-XS-core, VeRA-core, EVA-core, and random orthonormal bases. The ``-core'' suffix denotes our controlled fixed-basis instantiation, not the original method's full recipe; Appendix~\ref{app:baselines} gives the exact mappings. LoRA, DoRA, and PiSSA serve as trainable-basis rank-16 references.

\paragraph{Evaluation and selection.}
SVAMP and GSM8K use greedy decoding and exact match with a shared deterministic parser; the remaining tasks use per-choice conditional log likelihood. Appendix~\ref{app:scoring} motivates this task-format-specific choice and clarifies how the cross-task mean should be interpreted. We run seeds 42, 43, and 44 and report mean $\pm$ standard deviation. Held-out validation selects checkpoints and hyperparameters. \method{} fixes the learning rate at $5\times10^{-3}$ and searches warmup $\{0.03,0.1\}\times$ calibration size $\{128,256\}$; each matched-budget baseline searches learning rate $\{2,5\}\times10^{-3}$ at warmup 0.1. Thus baselines can tune learning rate, although \method{} evaluates more total warmup/calibration configurations. Table~\ref{tab:hyperparams} lists the complete protocol. The primary paired test is a two-sided Wilcoxon signed-rank test over the 11 unrounded task means.

\section{Results}
\input{tables/main_q3}

\subsection{Matched-budget performance at 3B}
Table~\ref{tab:main-q3} gives the complete Qwen2.5-3B comparison. \method{} reaches an 83.0 mean, 2.3 points above the next-best matched-budget constructor, LoRA-XS-core. It is best or tied on seven tasks and remains within 1.2 points of the best method on the other four. The largest gains over RawGrad occur on OpenBookQA (+10.0), HellaSwag (+7.4), and WinoGrande (+18.8), but the advantage is not carried by those tasks alone: every one of the 11 differences is positive.

Using unrounded task means, \method{} exceeds RawGrad by 4.78 points on average ($p=.001$) and every unadjusted Wilcoxon comparison against a matched-budget baseline is below $.05$. Table~\ref{tab:sig-main} summarizes the paired tests; Appendix~\ref{app:significance} reports win counts and paired-$t$ references. The task is the unit of replication, so these tests support a cross-task comparison and should not be read as significance claims for each individual dataset.

\begin{table}[htb]
\centering
\small
\setlength{\tabcolsep}{5pt}
\begin{tabular}{lrr}
\toprule
Baseline & Mean $\Delta$ & Wilcoxon $p$ \\
\midrule
RawGrad & +4.78 & .001 \\
FiLoRA-core & +3.51 & .002 \\
MiLoRA-core & +2.57 & $<.01$ \\
LoRA-XS-core & +2.31 & $<.01$ \\
VeRA-core & +2.74 & .014 \\
EVA-core & +3.09 & .007 \\
Random & +3.28 & $<.01$ \\
\bottomrule
\end{tabular}
\caption{FCCA minus each 3B baseline over 11 unrounded task means. Complete win counts, paired-$t$ references, and multiplicity caveats appear in Appendix~\ref{app:significance}.}
\label{tab:sig-main}
\end{table}

\emph{At the principal 3B setting, FCCA's advantage is broad across tasks and extends to every matched-budget constructor.} The matched-$r^2$ protocol is central to that interpretation. A trainable rank-$r$ factor can rotate during optimization, so its final accuracy conflates initialization quality with the ability to repair a weak starting span. Freezing $P,Q$ removes that repair mechanism and makes basis selection directly observable. Frozen-core performance therefore diagnoses span quality and models state-constrained adaptation; it does not imply that practical adapters must always freeze their bases.

The ordering is not a simple complexity ladder. VeRA and FiLoRA-core win isolated tasks, showing that several low-dimensional spans can be adequate locally; their lower cross-task means show that this adequacy does not transfer reliably. RawGrad is the decisive control: it uses exactly the same signed cross-moment as FCCA, so its deficit on all 11 tasks isolates the metric applied at truncation, not access to additional supervision. EVA-core remains competitive at larger scales, but its high 3B variance on MRPC and HellaSwag exposes task-level brittleness in this campaign.

The weight-spectrum controls sharpen the mismatch. LoRA-XS-core freezes the dominant singular directions of $W$, whereas MiLoRA-core emphasizes the minor spectrum; neither prior transfers consistently. LoRA-XS-core is the strongest 3B alternative, yet EVA-core is runner-up in the other model settings, and MiLoRA-core is competitive only on a subset of classification and QA tasks. The problem is therefore not merely choosing the ``wrong end'' of the spectrum: spectral energy describes how pretraining uses a matrix, whereas a downstream update depends on how task errors couple to current activations. A frozen basis cannot rotate a generic spectral prior into that coupling.

The data-derived baselines expose the complementary boundary. FiLoRA-core edges FCCA on SVAMP (80.1 versus 79.7), and EVA-core nearly ties it on Llama-1B; marginal statistics can therefore suffice when the update aligns with high-variance activation or curvature directions. Their failures are concentrated: at 3B, FiLoRA-core trails FCCA by 7.7 points on HellaSwag and 12.8 on WinoGrande, while EVA-core is lower and highly variable on HellaSwag. This pattern does not identify a unique causal mechanism, but it matches the information distinction in Table~\ref{tab:basis-signals}: marginals identify energetic directions, whereas $G=\bbE[\delta x^\top]$ retains their signed pairing with task errors. FCCA helps when both that pairing and the local metric matter before truncation.

\subsection{Scale, architecture, and task-family coverage}
\begin{figure*}[htb]
\centering
\begin{subfigure}[t]{0.240\textwidth}
  \centering\includegraphics[width=\linewidth]{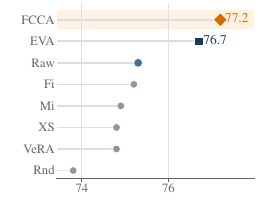}
  \caption{Qwen2.5-1.5B.}
  \label{fig:scale-overview-a}
\end{subfigure}%
\hfill
\begin{subfigure}[t]{0.240\textwidth}
  \centering\includegraphics[width=\linewidth]{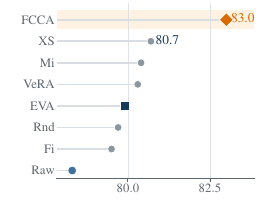}
  \caption{Qwen2.5-3B.}
  \label{fig:scale-overview-b}
\end{subfigure}%
\hfill
\begin{subfigure}[t]{0.240\textwidth}
  \centering\includegraphics[width=\linewidth]{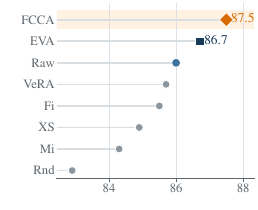}
  \caption{Qwen2.5-7B.}
  \label{fig:scale-overview-c}
\end{subfigure}%
\hfill
\begin{subfigure}[t]{0.240\textwidth}
  \centering\includegraphics[width=\linewidth]{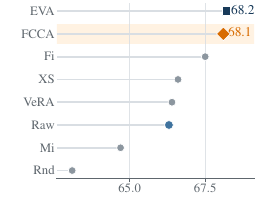}
  \caption{Llama-3.2-1B.}
  \label{fig:scale-overview-d}
\end{subfigure}
\caption{Cross-model matched-budget comparison. Each panel orders the eight frozen-core constructors by 11-task mean. FCCA is orange, RawGrad blue, EVA-core navy, and the remaining methods gray; Appendix~\ref{app:fullscale} reports exact means and task-level results.}
\label{fig:scale-overview}
\end{figure*}

Figure~\ref{fig:scale-overview} shows that the main result is not confined to one model size. \method{} ranks first on Qwen2.5-1.5B, 3B, and 7B, with means of 77.2, 83.0, and 87.5. Its lead over the strongest alternative is 0.5, 2.3, and 0.8 points, respectively, even though the identity of that alternative changes. At 7B, \method{} significantly exceeds six of seven controls using the supplied unrounded task means; EVA-core remains statistically comparable ($p=.054$). On Llama-3.2-1B, EVA-core leads by only 0.1 point (68.2 versus 68.1), while \method{} still exceeds RawGrad on eight tasks.

\takeaway{FCCA leads across all three Qwen scales and remains within 0.1 point of the best Llama result.} This boundary is informative. Larger backbones can expose several competitive low-dimensional spans, yet FCCA's margin over RawGrad remains positive in all four settings: using the displayed task means, it wins 38 of 44 task comparisons, ties one, and loses five. Because the two methods use the same signed cross-moment $G$, their repeated separation isolates the metric imposed before truncation, not additional task information. The margin peaks at 3B instead of growing monotonically; the evidence supports a robust geometric correction without implying a scaling law.

The breadth also holds across task families. On Qwen2.5-3B, FCCA averages 73.9 on the two math tasks, 89.6 on the five GLUE tasks, and 79.4 on the four reasoning/QA tasks. These are 0.5, 1.4, and 1.2 points above the strongest alternative within each family. The largest family-level gain over RawGrad is on reasoning/QA (+9.5), but the GLUE and math results rule out an explanation tied only to generated arithmetic answers. Nor is the ranking driven by unusually small seed variance: FCCA's median standard deviation is 1.0 point, versus 0.7--1.2 for the alternatives. Complete 1.5B, 7B, and Llama results are in Appendix~\ref{app:fullscale}.

\subsection{Whitening changes which directions survive truncation}
\begin{figure*}[htb]
\centering
\begin{subfigure}[t]{0.310\textwidth}
  \centering\includegraphics[width=\linewidth]{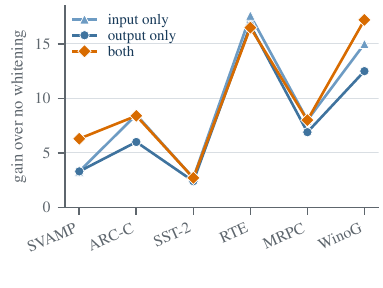}
  \caption{Whitening gains by task.}
  \label{fig:mechanism-a}
\end{subfigure}%
\hfill
\begin{subfigure}[t]{0.310\textwidth}
  \centering\includegraphics[width=\linewidth]{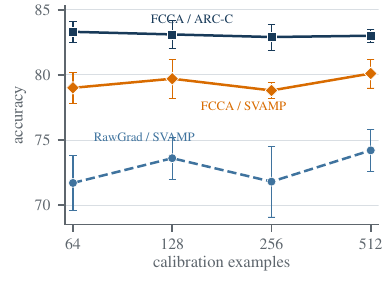}
  \caption{Accuracy versus calibration size.}
  \label{fig:mechanism-b}
\end{subfigure}%
\hfill
\begin{subfigure}[t]{0.310\textwidth}
  \centering\includegraphics[width=\linewidth]{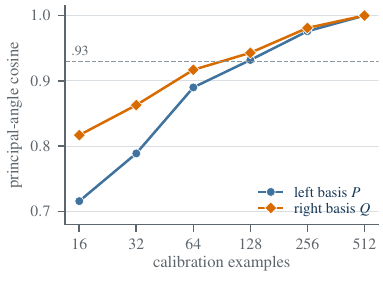}
  \caption{Basis convergence.}
  \label{fig:mechanism-c}
\end{subfigure}
\caption{Mechanism and calibration diagnostics on Qwen2.5-3B. (a) Gain from one- and two-sided whitening over the unwhitened SVD. (b) Accuracy versus calibration size (mean $\pm$ standard deviation, three seeds). (c) Mean principal-angle cosine to the 512-example bases. Appendix~\ref{app:ablations} reports exact values and additional controls.}
\label{fig:mechanism}
\end{figure*}

Figure~\ref{fig:mechanism}(a) isolates the paper's central intervention. \takeaway{Whitening improves every tested mechanism task, with two-sided gains of 2.7--17.2 points over the unscaled SVD.} The effect spans math, language understanding, and commonsense reasoning, and is largest on RTE and WinoGrande. Two-sided whitening is best or tied on four of six tasks; input-only is 0.1 point higher on ARC-C and 1.1 higher on RTE. Two-sided whitening is therefore the strongest overall default in this study, although input-only is slightly better on two individual tasks.

The comparison also separates task signal from local geometry. RawGrad preserves the signed activation--error pairing but imposes rank in Euclidean coordinates. EVA-core and FiLoRA-core encode aspects of the calibration distribution or marginal curvature without retaining that signed pairing. FCCA combines both exactly where information is discarded: before the rank-$r$ truncation. This ordering matters because whitening after truncation could rescale a chosen span but could not recover directions already removed. Appendix~\ref{app:proof} formalizes the same point through the local Fisher objective, while Appendix~\ref{app:ablations} reports the six-task values underlying the figure.

\subsection{Calibration saturates before the basis fully converges}
Figure~\ref{fig:mechanism}(b)--(c) distinguish downstream sufficiency from numerical convergence. Across 64--512 examples, FCCA varies by only 1.3 points on SVAMP and 0.4 on ARC-C, and it remains 5.9--7.3 points above RawGrad at every tested SVAMP size. The selected bases continue to move over the same range: their principal-angle cosines first exceed .93 on both sides at 128 examples and approach one only at 512. Accuracy therefore plateaus before the estimated span becomes identical to the largest-calibration reference.

A plausible interpretation is that several nearby rank-16 spans contain similarly useful directions and the square core can reweight coordinates within any of them. The mismatched-source experiment provides a sharper boundary. Calibrating SVAMP with GSM8K costs 1.7 points, whereas ARC-C and SST-2 calibration cost 5.1 and 7.1. Increasing sample count therefore cannot substitute for representative task signal. The ordering matches FCCA's statistic: calibration sets both marginal scales and the orientation of $\delta x^\top$. Related math data preserve more of this joint structure; additional out-of-domain examples can reduce variance around a mismatched population quantity without correcting its bias. Basis reuse therefore appears more plausible within task families or after a small target-task refresh. In the tested regime, \emph{coverage is more consequential than exact convergence to one reference basis}: modest, task-matched calibration is sufficient, but general cross-domain basis reuse is not established.

Principal-angle convergence and downstream utility therefore answer different questions. The former measures recovery of a particular reference basis; the latter only requires a span whose projected gradient and local geometry are adequate for the task. This distinction explains why accuracy can saturate before the basis stabilizes and cautions against choosing calibration size from geometric convergence alone.

\subsection{Accuracy--state trade-offs}
\begin{table}[htb]
\centering
\small
\setlength{\tabcolsep}{3.6pt}
\begin{tabular}{lrrr}
\toprule
Method & Mean & Trainable & $\Delta$ vs. FCCA \\
\midrule
\textbf{FCCA} & \textbf{83.0} & \textbf{36.9K} & reference \\
LoRA & 83.3 & 7.37M & +0.32 \\
DoRA & 83.2 & 7.54M & +0.23 \\
PiSSA & 80.4 & 7.37M & $-2.60$ \\
\bottomrule
\end{tabular}
\caption{Qwen2.5-3B accuracy and trainable state. Means are over the 11-task suite.}
\label{tab:efficiency-state}
\end{table}

\begin{table}[htb]
\centering
\scriptsize
\setlength{\tabcolsep}{2.25pt}
\begin{tabular}{lrrrr}
\toprule
Method & Init./calib. & Step & Projected total & Peak GPU \\
\midrule
\textbf{FCCA} & 35 s & 0.81 s & 4.1 min & 6.92 GB \\
LoRA & none & 0.92 s & 4.0 min & 7.19 GB \\
DoRA & none & 1.31 s & 5.7 min & 7.27 GB \\
PiSSA & 41 s & 1.76 s & 8.4 min & 7.20 GB \\
\bottomrule
\end{tabular}
\caption{SVAMP phase profile on one H20; totals project approximately 262 optimizer steps. The first column reports FCCA calibration and PiSSA initialization; PiSSA uses a one-time weight-SVD to initialize its trainable factors.}
\label{tab:efficiency-runtime}
\end{table}

Table~\ref{tab:efficiency-state} asks how much accuracy is lost when the selected span cannot move. FCCA's 83.0 mean is 0.318 below LoRA and 0.227 below DoRA despite optimizing 36,864 rather than 7.37--7.54M parameters. It exceeds LoRA on four tasks and DoRA on five; the trainable factors retain their clearest advantages on OpenBookQA and ARC-C. \takeaway{The result is near-parity at a $200\times$ reduction in trainable state.} The complete task-level comparison is in Appendix~\ref{app:trainable}.

The ordering clarifies what a frozen span gives up. LoRA and DoRA can rotate their factors throughout training, so they can repair calibration directions that are incomplete or task-locally misaligned. Their largest gains on OpenBookQA and ARC-C are consistent with a greater need for span revision on those tasks, although this comparison alone cannot establish the cause. PiSSA starts from dominant weight directions and then learns both factors; its lower 3B mean, driven especially by GSM8K, shows that a favorable weight-space initialization is not sufficient when the downstream update is poorly aligned with the pretrained spectrum. In contrast, FCCA spends its calibration budget on the downstream loss and relies on the square core only to recombine selected directions.

Table~\ref{tab:efficiency-runtime} places the state reduction in a runtime context. FCCA adds a one-time 35-second calibration pass, but its core-only updates yield a projected 4.1-minute run: close to LoRA (4.0), below DoRA (5.7), and roughly half PiSSA (8.4). The GPU-memory difference is modest because frozen model weights dominate at this scale; the more substantial savings are in gradients, optimizer state, and communication. The storage claim is narrower: before merging, task-specific $P,Q$ bases remain width dependent, and after merging deployment requires a task-specific model copy. Appendix~\ref{app:repro} gives the full protocol.

Under these measurements, FCCA amortizes its calibration relative to DoRA after roughly 70 optimizer steps and relative to LoRA after roughly 318. The 262-step SVAMP run is therefore well beyond the DoRA break-even point and just below the LoRA break-even point. These thresholds are hardware- and implementation-specific, but they make the runtime trade-off explicit.

\subsection{Span selection and coordinate conditioning are distinct}
\begin{table}[htb]
\centering
\small
\setlength{\tabcolsep}{3.0pt}
\renewcommand{\arraystretch}{1.04}
\begin{tabular}{lccc}
\toprule
Setting & SVAMP & ARC-C & GSM8K \\
\midrule
\multicolumn{4}{l}{\textbf{Span size} \hspace{0.35em}\textit{(all $q/k/v/o$)}} \\
$r=8$  & 73.9$\pm$0.3 & \textbf{84.1$\pm$0.2} & 67.1$\pm$2.6 \\
$r=16$ & \textbf{79.7$\pm$1.5} & 83.1$\pm$1.1 & \textbf{67.3$\pm$1.2} \\
$r=32$ & 74.9$\pm$2.1 & 81.3$\pm$1.6 & 63.4$\pm$1.2 \\
$r=64$ & 70.6$\pm$0.9 & 69.8$\pm$1.4 & 56.7$\pm$1.7 \\
\addlinespace[1pt]
\multicolumn{4}{l}{\textbf{Core coordinates} \hspace{0.35em}\textit{(fixed span)}} \\
w/ thin QR & \textbf{79.7$\pm$1.5} & \textbf{83.1$\pm$1.1} & \textbf{69.3$\pm$1.7} \\
w/o QR$^{\dagger}$ & div. (0.0) & 22.9$\pm$1.9 & div. (0.0) \\
\addlinespace[1pt]
\multicolumn{4}{l}{\textbf{Whitening estimator} \hspace{0.35em}\textit{(128--256 calibration)}} \\
Diagonal & \textbf{79.7$\pm$1.5} & \textbf{83.1$\pm$1.1} & \textbf{69.3$\pm$1.7} \\
Full K-FAC & 71.0$\pm$2.6 & 82.7$\pm$1.6 & 67.6$\pm$2.2 \\
\bottomrule
\end{tabular}
\caption{Design diagnostics on Qwen2.5-3B. $^{\dagger}$Without QR, SVAMP and GSM8K diverged under the shared learning rate (loss $\rightarrow\infty$); 0.0 is the resulting exact-match score, not a converged accuracy. GSM8K uses the dedicated ablation configuration (QR reference 69.3).}
\label{tab:design-summary}
\end{table}

The span-size block of Table~\ref{tab:design-summary} shows a consistent small-rank regime and a clear breakdown at larger ranks. Performance peaks at $r=16$ on SVAMP and GSM8K, while ARC-C is already strongest at $r=8$; all three tasks drop substantially at $r=32$ and again at $r=64$. This pattern is not an expressivity paradox: for a fixed calibration statistic, the top-$r$ SVD spans are nested, so a larger span can reproduce the lower-rank solution. What changes is the finite-sample estimation and optimization problem. Ranks above 16 admit weaker tail directions estimated from the same 128--256 calibration examples, while the core grows from $16^2=256$ to $32^2=1024$ and $64^2=4096$ coefficients under essentially fixed training data, steps, and learning-rate search. QR removes gross anisotropy, but it does not denoise tail directions or regularize the larger core. The evidence therefore supports a small-to-moderate rank sweet spot under this protocol, with $r=8$--$16$ covering the best tested points across tasks.

\begin{table}[htb]
\centering
\scriptsize
\setlength{\tabcolsep}{3.2pt}
\begin{tabular}{lrrr}
\toprule
Configuration & FCCA & FiLoRA-c & RawGrad \\
\midrule
$r=8$, $q/v$ & 74.7$\pm$2.1 & \textbf{75.4$\pm$0.8} & 75.1$\pm$0.2 \\
$r=8$, $q/k/v/o$ & 75.2$\pm$1.3 & \textbf{76.8$\pm$1.1} & 74.2$\pm$0.4 \\
$r=16$, $q/v$ & 76.0$\pm$2.8 & \textbf{77.3$\pm$1.2} & 76.9$\pm$0.3 \\
$r=16$, $q/k/v/o$ & 79.7$\pm$1.5 & \textbf{80.1$\pm$0.4} & 74.9$\pm$2.0 \\
\bottomrule
\end{tabular}
\caption{SVAMP rank-by-target-module interaction. FCCA's separation from RawGrad appears at the main $r=16$, $q/k/v/o$ budget; FiLoRA-core remains slightly higher in that cell.}
\label{tab:rank-modules-main}
\end{table}

Table~\ref{tab:rank-modules-main} sharpens this boundary. FCCA is near RawGrad in the three smaller configurations and opens a 4.8-point gap only at $r=16$, $q/k/v/o$; FiLoRA-core remains 0.4 point higher there. Fisher scaling therefore does not win every small-budget cell. Its clearest benefit appears when rank and module coverage are sufficient to retain the balanced directions selected by the whitened cross-covariance.

Taken together, the two budget studies favor distributing a moderate rank across all four attention projections. Broader module coverage exposes more types of task-relevant transformations, whereas increasing $r$ beyond 16 mainly adds weaker tail directions under the same calibration and optimization budget. In these campaigns, breadth across $q/k/v/o$ is more useful than indiscriminately expanding each frozen span.

The coordinate block isolates optimization geometry. QR preserves the update family, yet without it the shared-learning-rate runs diverge on SVAMP and GSM8K and reach 22.9 on ARC-C. Thus the 0.0 entries diagnose failed optimization in the raw coordinates, not zero span expressivity. The estimator block isolates finite-sample curvature estimation: dense K-FAC must infer two width-by-width covariances from only 128--256 examples and trails the diagonal estimator on all three tasks. This supports diagonal whitening in the tested low-calibration regime, while leaving open whether stronger shrinkage, structured factors, or substantially more calibration could reverse the result.

\takeaway{A useful frozen adapter requires both a well-chosen span and a trainable coordinate system.} The controls separate three failure sources that end-task accuracy alone would conflate: the statistic can omit task information, the rank metric can discard locally useful directions, or the coordinates can make an adequate span untrainable.

The controls differ in inferential strength. Whitening-side and QR ablations hold the pipeline fixed, and QR preserves the update family; dense K-FAC and rank sweeps also change finite-sample conditioning or model size. We therefore interpret the latter as regime-specific design evidence, not monotonic laws.

\section{Conclusion and Discussion}
FCCA combines signed task signal, Fisher-scaled truncation, and QR-conditioned core coordinates. It leads at all Qwen scales, nearly ties on Llama-1B, and approaches LoRA and DoRA with roughly $200\times$ less trainable state while keeping calibration and basis-storage costs explicit.

The baseline pattern separates three design choices. Weight-spectrum and random bases lack downstream error information; EVA-core and FiLoRA-core retain useful marginals but omit the signed input--error pairing; RawGrad preserves that pairing but imposes rank in an unscaled metric; and no-QR keeps the selected span yet makes its core effectively untrainable. Trainable factors retain one advantage: they can rotate to repair an incomplete span, which explains the remaining gap to LoRA and DoRA.

Frozen-core accuracy is therefore a subspace diagnostic. Under resource constraints, the practical lesson is to preserve signed task signal in the metric where rank is imposed, allocate moderate rank before admitting weak tail directions, and expose the retained span through trainable coordinates.
\FloatBarrier
\section*{Limitations}
We evaluate fixed subsets (300 examples per math task and 500 per remaining task), so absolute scores are not directly comparable with full-test leaderboards. The heterogeneous 11-task macro-average weights tasks equally, and the paired Wilcoxon test has only 11 observations.

The study uses three seeds, one software stack, and one GPU family. Three Qwen scales are complemented by one Llama setting, and several ablations cover only one to three tasks; their design conclusions may be regime specific. We do not evaluate instruction following or open-ended generation beyond exact-match math.

Calibration depends on task match and requires a backward pass plus cross-covariances that can consume host memory. The $r^2$ budget measures trainable and optimizer state, not complete storage. Search trial counts also differ (four FCCA warmup/calibration configurations versus two baseline learning rates), and benchmark contamination was not audited.

\section*{Acknowledgments}
Supported by the Noncommunicable Chronic Diseases-National Science and Technology Major Project (2026ZD0558202) and the Ningbo Yongjiang Talent Introduction Programme (2023A-399-G).

\clearpage
\bibliography{custom}

\newpage
\appendix
\section{Mathematical Details and Coordinate Invariance}
\label{app:proof}

\subsection{Projected-gradient criterion for a frozen core}
Let $\phi(R)=\mathcal L(W+sPRQ^\top)$ for fixed $P$ and $Q$, and let $G=\nabla_W\mathcal L(W)$. The first-order expansion around the zero core is
\begin{align}
\phi(R)
&=\phi(0)+s\langle G,PRQ^\top\rangle+o(\|R\|_F)\\
&=\phi(0)+s\langle P^\top GQ,R\rangle+o(\|R\|_F).
\end{align}
Hence $\nabla_R\phi(0)=sP^\top GQ$. A gradient step $R_1=-\gamma sP^\top GQ$ changes the linearized objective by $-\gamma s^2\|P^\top GQ\|_F^2$. The fixed scale $s$ changes the magnitude of the step but not the ranking of candidate subspaces. This derivation motivates the matched-budget comparison: a useful constructor must capture the signed task cross term, not only high-variance or high-weight-energy directions.

\subsection{Proof of Theorem 1}
Set $Z=D^{1/2}\Delta W A^{1/2}$. Positive definiteness makes the change of variables invertible, gives $\Delta W=D^{-1/2}ZA^{-1/2}$, and preserves rank. Because the inverse square roots are symmetric,
\begin{align}
\langle G,\Delta W\rangle
&=\Tr\!\left(G^\top D^{-1/2}ZA^{-1/2}\right)\\
&=\Tr\!\left((D^{-1/2}GA^{-1/2})^\top Z\right)\\
&=\langle\tgrad,Z\rangle.
\end{align}
Substituting into Eq.~\eqref{eq:local} and completing the square yields
\begin{equation}
\min_{\rank(Z)\le r}
\frac{1}{2\eta}\|Z+\eta\tgrad\|_F^2
-\frac{\eta}{2}\|\tgrad\|_F^2.
\end{equation}
The second term is constant in $Z$. By the Eckart--Young--Mirsky theorem, the best rank-$r$ approximation to $-\eta\tgrad$ in Frobenius norm is $-\eta[\tgrad]_r$. Mapping this solution back through the inverse square roots gives Eq.~\eqref{eq:solution}.

\paragraph{Why truncation is performed after whitening.}
Except when $A$ and $D$ are scalar multiples of the identity (or share special symmetries with $G$), the leading singular vectors of $D^{-1/2}GA^{-1/2}$ are not obtained by simply rescaling the leading singular vectors of $G$. Whitening can change both singular directions and their ordering. FCCA therefore applies the rank constraint in Fisher coordinates before any raw-gradient directions are discarded.

\subsection{Thin QR leaves the update family unchanged}
Let $P_0=PT_P$ and $Q_0=QT_Q$ be thin QR decompositions. If the selected columns have full rank, the triangular factors $T_P,T_Q\in\R^{r\times r}$ are nonsingular and
\begin{equation}
\{P_0ZQ_0^\top:Z\in\R^{r\times r}\}
=
\{PRQ^\top:R\in\R^{r\times r}\},
\end{equation}
with the bijection $R=T_PZT_Q^\top$. QR therefore changes optimization coordinates, not expressivity. Zero initialization is function preserving in either coordinate system, since $Z=0$ if and only if $R=0$.

\subsection{Diagonal moments, damping, and terminology}
For each target matrix, calibration accumulates the uncentered cross moment $G$ and coordinatewise second moments $a=\bbE[x^{\odot2}]$ and $d=\bbE[\delta^{\odot2}]$. The default factors are $A=\diag(a)+\epsilon I$ and $D=\diag(d)+\epsilon I$, where damping keeps inverse square roots finite in rarely activated coordinates. The term ``cross-covariance'' follows common gradient-factor terminology; no mean subtraction is applied because $\bbE[\delta x^\top]$ is the empirical loss gradient of the linear map. Dense $A$ and $D$ are evaluated only in the dedicated full K-FAC ablation.

\section{Complete Experimental Protocol}
\label{app:protocol}

\subsection{Datasets, fixed subsets, and training length}
Table~\ref{tab:datasets} gives the complete data protocol, including source identifiers, fixed evaluation subsets, and training lengths.
\begin{table*}[htb]
\centering
\scriptsize
\setlength{\tabcolsep}{4.2pt}
\begin{tabular}{lllrrr}
\toprule
Task & Dataset source & Configuration & Evaluation & Train & Eval / epochs \\
\midrule
SVAMP & \texttt{ChilleD/SVAMP} & default & greedy exact match & 700 & 300 / 3 \\
GSM8K & \texttt{openai/gsm8k} & \texttt{main} & greedy exact match & 1500 & 300 / 2 \\
SST-2 & \texttt{nyu-mll/glue} & \texttt{sst2} & likelihood MC & 2000 & 500 / 1 \\
QNLI & \texttt{nyu-mll/glue} & \texttt{qnli} & likelihood MC & 2000 & 500 / 1 \\
CoLA & \texttt{nyu-mll/glue} & \texttt{cola} & likelihood MC & 2000 & 500 / 2 \\
RTE & \texttt{nyu-mll/glue} & \texttt{rte} & likelihood MC & 2000 & 500 / 3 \\
MRPC & \texttt{nyu-mll/glue} & \texttt{mrpc} & likelihood MC & 2000 & 500 / 3 \\
ARC-Challenge & \texttt{allenai/ai2\_arc} & \texttt{ARC-Challenge} & likelihood MC & 1119 & 500 / 3 \\
OpenBookQA & \texttt{allenai/openbookqa} & \texttt{main} & likelihood MC & 2000 & 500 / 3 \\
HellaSwag & \texttt{Rowan/hellaswag} & default & likelihood MC & 2000 & 500 / 2 \\
WinoGrande & \texttt{allenai/winogrande} & \texttt{winogrande\_xl} & likelihood MC & 2000 & 500 / 2 \\
\bottomrule
\end{tabular}
\caption{Complete dataset protocol. ``Default'' denotes a dataset source with no named Hugging Face configuration. ``Eval / epochs'' gives the fixed evaluation-subset size and training epochs. All datasets were loaded from an offline cache, and the same fixed subsets were reused for every method and seed.}
\label{tab:datasets}
\end{table*}

The campaign uses a common evaluation budget instead of each benchmark's full official test set: 300 examples for each math task and 500 for every other task. This keeps the four-model, eight-method, three-seed comparison tractable and ensures that every method is evaluated on identical examples. It also means the absolute scores should be interpreted within this protocol rather than compared directly with full-test leaderboard numbers.

\subsection{Scoring protocols and aggregate interpretation}
\label{app:scoring}
For SVAMP and GSM8K, decoding is greedy. The parser takes the first occurrence of either ``The answer is:'' or ``\#\#\#\#'' and then applies one deterministic numeric normalizer shared by all methods. Taking the first marker prevents a model that repeats or revises an answer later in the generation from being scored under a different extraction rule.

For the remaining nine tasks, each candidate answer is scored by its conditional log likelihood under the same prompt and the highest-scoring choice is selected. These results do not depend on generation temperature, stopping rules, or label-token parsing. CoLA and the other GLUE tasks are therefore reported as accuracy-like choice scores in this campaign, not necessarily with each benchmark's official leaderboard metric.

\paragraph{Why decoding and likelihood are combined.}
The two scoring modes follow task format, not method choice. Mathematical reasoning requires a generated numeric answer, whereas conditional likelihood avoids label-token and stopping-rule artifacts for classification and multiple choice. The 11-task mean is consequently an unweighted macro-average of heterogeneous accuracy-like scores on a common $0$--$100$ scale. We use it as a descriptive summary, not a calibrated utility measure, and accompany it with per-task results, family means, and rank-based paired tests.

\subsection{Common adapter and hyperparameter protocol}
Table~\ref{tab:hyperparams} summarizes the unified optimization protocol.
\begin{table*}[htb]
\centering
\small
\setlength{\tabcolsep}{5pt}
\begin{tabular}{p{3.0cm}p{4.3cm}p{4.3cm}}
\toprule
Component & FCCA & Matched-budget baselines \\
\midrule
Shared adapter & rank 16; all $q/k/v/o$ projections; every layer; zero $r\times r$ core & same \\
Precision / batch & bf16; batch size 1; gradient accumulation 8 (effective batch 8) & same \\
Trainable budget & one $r\times r$ core per target matrix; 36,864 parameters at Qwen2.5-3B & same \\
Learning rate & fixed $5\times10^{-3}$ & validation search over $\{2,5\}\times10^{-3}$ \\
Warmup & validation search over $\{0.03,0.1\}$ & fixed 0.1 \\
Calibration size & validation search over $\{128,256\}$ & method-specific basis construction under the common data budget \\
Final evaluation & seeds 42, 43, 44; mean $\pm$ sample standard deviation & same \\
\bottomrule
\end{tabular}
\caption{Unified optimization protocol. FCCA has a single fixed learning rate and searches its own warmup/calibration knobs; matched-budget baselines can select learning rate.}
\label{tab:hyperparams}
\end{table*}

FCCA evaluates four warmup-by-calibration combinations at one learning rate. Each matched-budget baseline evaluates two learning rates at fixed warmup. This gives baselines an LR-selection advantage but does not equalize total search trials: FCCA has four configurations and a baseline has two. The best configuration is selected by held-out validation before the three-seed evaluation. LoRA, DoRA, and PiSSA are separate trainable-basis references and use validation-selected learning rates appropriate to their parameterization.

\subsection{Calibration implementation}
Calibration examples are drawn only from the task's training split. With \texttt{calib\_group\_size=999}, one grouped hooked forward/backward pass collects statistics for all target layers. Cross moments and diagonal second moments are placed on CPU (\texttt{stats\_device=cpu}); after the SVD and QR steps, the statistics are discarded and only the bases and cores remain. Diagonal whitening is the default. Full dense K-FAC factors are constructed only for the controlled comparison in Table~\ref{tab:design-summary}.

\subsection{Statistical reporting}
Every displayed cell is mean $\pm$ sample standard deviation over seeds 42, 43, and 44 unless explicitly marked as a mean-only mechanism table. The primary cross-task test is a two-sided paired Wilcoxon signed-rank test over the 11 unrounded task means. A paired $t$-test is shown as a reference, but the rank test is primary because the 11 task differences include large outliers and provide little basis for a normality assumption. The task, not the seed, is the unit of replication for this analysis.

\section{Baseline Constructions}
\label{app:baselines}
All matched-budget baselines are converted to the same frozen-basis, trainable-core architecture. Consequently, labels ending in ``-core'' denote controlled instantiations of a basis-selection idea, not a reproduction of the original method's complete parameterization or native trainable budget.

\paragraph{RawGrad.}
Compute $G=U\Sigma V^\top$ on the calibration set and take $P=U_r$, $Q=V_r$. No input- or output-side whitening is applied. RawGrad therefore has the same signed task cross moment as FCCA and is the cleanest control for the effect of $A$ and $D$.

\paragraph{FiLoRA-core.}
Select fixed input and output bases from the marginal K-FAC/Fisher factors $A$ and $D$. This retains curvature-aware marginal eigenspaces but does not preserve the signed input--error pairing in $G$.

\paragraph{MiLoRA-core and LoRA-XS-core.}
Both use the pretrained weight SVD and are task agnostic during basis construction. MiLoRA-core selects the bottom-$r$ weight singular directions. LoRA-XS-core uses the top-$r$ directions and incorporates the singular values into the left basis before the common coordinate handling and core training.

\paragraph{EVA-core.}
The input-side basis is obtained from the leading singular directions of calibration activations, following the explained-variance principle. The campaign uses the corresponding fixed output-side construction from its EVA-core implementation and trains only the common $r\times r$ core.

\paragraph{VeRA-core and Random-core.}
VeRA-core preserves the frozen-random-basis principle but replaces VeRA's native lightweight scaling parameterization with the common square core. Random-core samples independent orthonormal left and right bases. Their purpose is to separate the value of task-derived structure from the capacity of the shared core itself.

\paragraph{Trainable-basis references.}
LoRA, DoRA, and PiSSA update rank-$r$ factors of width-dependent size, approximately $r(m+n)$ values per target matrix. PiSSA initializes those trainable factors from principal weight components; unlike LoRA-XS-core, its factors remain trainable. These references answer how much accuracy is lost by freezing a carefully selected subspace, not which method wins under equal trainable parameters.

\section{Full Cross-Scale Results}
\label{app:fullscale}
Figure~\ref{fig:scale-overview} emphasizes the ranking structure; Table~\ref{tab:cross-scale} gives the exact 11-task means. Tables~\ref{tab:full-q15}, \ref{tab:full-q7}, and~\ref{tab:full-llama1} report every task-level mean and standard deviation for the three non-primary settings, while the Qwen2.5-3B campaign remains in Table~\ref{tab:main-q3}.
\input{tables/cross_scale}
\input{tables/full_q15}
\input{tables/full_q7}
\input{tables/full_llama1}

On Qwen2.5-1.5B, FCCA has the highest displayed 11-task mean (77.15 before rounding), 0.47 point above EVA-core and 1.90 above RawGrad. On Qwen2.5-7B, FCCA reaches 87.49, 0.79 above EVA-core; the supplied unrounded Wilcoxon comparison to EVA-core is $p=.054$, so the two are statistically comparable at that scale. On Llama-3.2-1B, EVA-core leads by 0.13 point before rounding, while FCCA exceeds RawGrad on eight of 11 tasks. These results support a repeated Qwen scaling pattern and a top-tier cross-architecture result, rather than a claim that one constructor must win on every model family.

All averages in Table~\ref{tab:cross-scale} are recomputed from the 11 displayed task means. This avoids silently changing the task set across models and keeps the comparison aligned with the primary 3B campaign. The main-text figure highlights FCCA and RawGrad because they isolate the effect of Fisher scaling while sharing the same signed cross-moment; the complete tables retain all six additional constructors.

\section{Trainable-Basis References}
\label{app:trainable}
\input{tables/trainable_basis_full}

LoRA, DoRA, and PiSSA train both rank-16 factors rather than a square core. On Qwen2.5-3B, FCCA is 0.318 point below LoRA and 0.227 below DoRA on the 11-task mean despite training about $200\times$ fewer parameters. FCCA is higher than LoRA on SVAMP, GSM8K, SST-2, and QNLI, and higher than DoRA on those four tasks plus WinoGrande. The largest trainable-basis advantages are on OpenBookQA and ARC-C. The comparison therefore supports near-parity on average, not dominance over full trainable factors.

An earlier Qwen2.5-7B PiSSA GSM8K run at learning rate $10^{-3}$ produced 0.5\% exact match, validation loss 5.48, and degenerate numeric strings. This was an optimizer divergence caused by an ill-matched learning rate, not evidence that PiSSA's subspace is intrinsically defective. Re-running at $2\times10^{-4}$ gives 77.2$\pm$2.1. The Qwen2.5-3B PiSSA value in Table~\ref{tab:trainable-basis-full} is a separate result: both candidate learning rates train stably, but GSM8K remains 51.4$\pm$1.7.

\section{Task-Level Significance}
\label{app:significance}
Table~\ref{tab:significance} expands the compact main-text significance summary with win counts and paired-$t$ references.
\begin{table}[htb]
\centering
\scriptsize
\setlength{\tabcolsep}{2.4pt}
\begin{tabular}{lrrrr}
\toprule
Baseline & Mean $\Delta$ & Wins & Wilcox. $p$ & $t$-test $p$ \\
\midrule
RawGrad & +4.78 & 11/11 & .001 & .015 \\
FiLoRA-core & +3.51 & 10/11 & .002 & .016 \\
MiLoRA-core & +2.57 & 10/11 & $<.01$ & .005 \\
LoRA-XS-core & +2.31 & 10/11 & $<.01$ & .017 \\
VeRA-core & +2.74 & 9/11 & .014 & .015 \\
EVA-core & +3.09 & 10/11 & .007 & .090 \\
Random & +3.28 & 11/11 & $<.01$ & $<.01$ \\
\bottomrule
\end{tabular}
\caption{FCCA minus each matched-budget baseline on Qwen2.5-3B, paired over 11 unrounded task means. ``Wins'' counts tasks on which FCCA is higher; rounded main-table cells can display a tie even when the unrounded comparison is nonzero.}
\label{tab:significance}
\end{table}

All reported tests are unadjusted. A Bonferroni threshold for seven comparisons would be $.05/7\approx.0071$, but several supplied values are available only after rounding or as bounds such as $p<.01$, so the corrected decisions cannot be audited exactly from the summary record. We therefore report the pre-specified unadjusted comparisons transparently and do not make a multiplicity-corrected significance claim.

The paired-$t$ reference for EVA-core is .090 even though the Wilcoxon value is .007. Two especially large positive FCCA margins (HellaSwag and MRPC) inflate the variance of the raw differences; the rank-based test is less sensitive to their magnitude. With only 11 task pairs, both the resolution and power of any cross-task test remain limited.

\section{Ablations and Robustness}
\label{app:ablations}

\subsection{Whitening sides}
Figure~\ref{fig:mechanism}(a) visualizes the gains, while Table~\ref{tab:whitening-full} preserves the underlying six-task means. Table~\ref{tab:side-svamp} is the strict exact-pipeline control in which learning rate, warmup, calibration size, and all other choices are fixed and only the whitening gates change.

\begin{table*}[htb]
\centering
\small
\setlength{\tabcolsep}{7.5pt}
\begin{tabular}{lrrrrr}
\toprule
Task & None: $\operatorname{svd}(G)$ & Input only & Output only & Both (FCCA) & Both$-$None \\
\midrule
SVAMP & 73.4 & 76.7 & 76.7 & \textbf{79.7} & +6.3 \\
ARC-C & 74.7 & \textbf{83.2} & 80.7 & 83.1 & +8.4 \\
SST-2 & 92.4 & \textbf{95.1} & 94.8 & \textbf{95.1} & +2.7 \\
RTE & 71.7 & \textbf{89.3} & 88.4 & 88.2 & +16.5 \\
MRPC & 80.4 & \textbf{88.4} & 87.3 & \textbf{88.4} & +8.0 \\
WinoGrande & 55.6 & 70.6 & 68.1 & \textbf{72.8} & +17.2 \\
\bottomrule
\end{tabular}
\caption{Whitening-side ablation on Qwen2.5-3B (three-seed means, $\times100$). Input/output divide by $\sqrt A$/$\sqrt D$; Both is FCCA. The study reports means where per-seed standard deviations were not retained.}
\label{tab:whitening-full}
\end{table*}

\begin{table}[htb]
\centering
\small
\begin{tabular}{lr}
\toprule
Configuration & SVAMP \\
\midrule
None: $\operatorname{svd}(G)$ & 73.4$\pm$2.1 \\
Input only: $GA^{-1/2}$ & 76.7$\pm$1.4 \\
Output only: $D^{-1/2}G$ & 76.7$\pm$2.9 \\
Both: $D^{-1/2}GA^{-1/2}$ & \textbf{79.7$\pm$1.5} \\
\bottomrule
\end{tabular}
\caption{Exact-pipeline whitening-side ablation on SVAMP. Only the whitening gates change.}
\label{tab:side-svamp}
\end{table}

Every whitened variant exceeds the no-whitening SVD on every populated task, and full whitening gains 2.7--17.2 points over None. Full whitening is best or tied on four of six tasks; input-only is 0.1 higher on ARC-C and 1.1 higher on RTE. The exact SVAMP control shows the same ordering under fixed non-whitening hyperparameters, strengthening the causal interpretation. The evidence supports geometry-aware normalization and two-sided whitening as a robust default, not a guarantee that adding the second side must improve finite-sample validation accuracy.

\subsection{Calibration size and basis stability}
Tables~\ref{tab:calib-accuracy} and~\ref{tab:basis-stability} separate downstream accuracy from convergence of the selected subspaces.
\begin{table}[htb]
\centering
\scriptsize
\setlength{\tabcolsep}{3.0pt}
\begin{tabular}{llrrrr}
\toprule
Method & Task & 64 & 128 & 256 & 512 \\
\midrule
FCCA & SVAMP & 79.0$\pm$1.2 & 79.7$\pm$1.5 & 78.8$\pm$0.6 & 80.1$\pm$1.1 \\
RawGrad & SVAMP & 71.7$\pm$2.1 & 73.6$\pm$1.6 & 71.8$\pm$2.7 & 74.2$\pm$1.6 \\
FCCA & ARC-C & 83.3$\pm$0.8 & 83.1$\pm$1.1 & 82.9$\pm$1.0 & 83.0$\pm$0.5 \\
\bottomrule
\end{tabular}
\caption{Accuracy sensitivity to calibration size (three seeds, $\times100$).}
\label{tab:calib-accuracy}
\end{table}

\begin{table}[htb]
\centering
\scriptsize
\setlength{\tabcolsep}{2.3pt}
\begin{tabular}{lrrrrrr}
\toprule
Calib. examples & 16 & 32 & 64 & 128 & 256 & 512 \\
\midrule
$\cos(P,P_{512})$ & .716 & .789 & .890 & .932 & .976 & 1.000 \\
$\cos(Q,Q_{512})$ & .817 & .863 & .917 & .943 & .981 & 1.000 \\
\bottomrule
\end{tabular}
\caption{Mean principal-angle cosine between the selected bases and the 512-example reference.}
\label{tab:basis-stability}
\end{table}

As Tables~\ref{tab:calib-accuracy} and~\ref{tab:basis-stability} show, accuracy is nearly flat from 64 to 512 examples even though the selected subspaces continue to converge. On SVAMP, FCCA spans 78.8--80.1 and remains 5.9--7.3 points above RawGrad at every calibration size; ARC-C spans only 0.4 point. The principal-angle cosines exceed .93 on both sides at 128 examples. These results show that the leading useful directions are identifiable with modest calibration data in the tested tasks, while not implying that all layers or domains will have the same sample requirement.

\subsection{Mismatched calibration}
\begin{table}[htb]
\centering
\small
\begin{tabular}{lrr}
\toprule
Calibration source & SVAMP & $\Delta$ from matched \\
\midrule
SVAMP (matched) & 79.7$\pm$1.5 & 0.0 (ref.) \\
GSM8K & 78.0$\pm$0.5 & $-1.7$ \\
ARC-C & 74.6$\pm$1.6 & $-5.1$ \\
SST-2 & 72.6$\pm$2.3 & $-7.1$ \\
\bottomrule
\end{tabular}
\caption{Training on SVAMP with bases calibrated on another dataset.}
\label{tab:mismatch}
\end{table}

Table~\ref{tab:mismatch} shows that same-domain GSM8K calibration loses only 1.7 points, whereas science QA and sentiment calibration lose 5.1 and 7.1. The most distant mismatch (72.6) is 0.8 point below the no-whitening SVAMP control (73.4), so mismatched calibration degrades gracefully here without guaranteeing an advantage over RawGrad. The ordering supports the task-aware interpretation: representative calibration data are what make the whitened cross moment especially useful.

\subsection{Rank and adapted modules}
Main-text Table~\ref{tab:rank-modules-main} separates the SVAMP rank-by-module grid from the dedicated all-projection sweep in Table~\ref{tab:design-summary}. The full sweep favors $r=16$ on SVAMP and GSM8K and $r=8$ on ARC-C; all three tasks deteriorate at $r=32$ and $r=64$. The sweep cannot isolate estimation noise, optimization difficulty, and overcapacity. Because the two tables come from separate controlled campaigns, their $r=8$, $q/k/v/o$ cells are not duplicate measurements.

\subsection{QR coordinates and dense K-FAC}
Main-text Table~\ref{tab:design-summary} reports the full three-seed QR and whitening-estimator controls. Without QR, raw inverse-whitened bases differ in scale by orders of magnitude: SVAMP and GSM8K diverge to infinite loss and zero exact match, while ARC-C falls to 22.9. QR adds no expressive directions, but it makes a common core optimizer usable in these runs.

\paragraph{Retuned no-QR control.}
We reran the no-QR variant at learning rates $\{5\times10^{-3},2\times10^{-3},10^{-3},5\times10^{-4}\}$, holding all other settings fixed. Table~\ref{tab:noqr-lr-sweep} reports the best observed score over the grid. SVAMP and GSM8K remain at 0.0 for every rate because the loss diverges or becomes NaN and generation produces no extractable answer. ARC-C peaks at 22.7, close to four-choice chance (25\%) and effectively unchanged from the 22.9 shared-learning-rate control.

\begin{table}[htb]
\centering
\scriptsize
\setlength{\tabcolsep}{5.0pt}
\begin{tabular}{lcl}
\toprule
Task & Best no-QR & Sweep outcome \\
\midrule
SVAMP & 0.0 & all four LRs diverged/NaN \\
ARC-C & 22.7 & near chance (25\%) \\
GSM8K & 0.0 & all four LRs diverged/NaN \\
\bottomrule
\end{tabular}
\caption{No-QR learning-rate sweep on Qwen2.5-3B ($r=16$, all $q/k/v/o$ projections). The grid is $\{5\times10^{-3},2\times10^{-3},10^{-3},5\times10^{-4}\}$; the table reports the best score observed for each task.}
\label{tab:noqr-lr-sweep}
\end{table}

Reducing the learning rate by an order of magnitude does not rescue the raw inverse-whitened coordinates. Since thin QR preserves the representable update family, the sweep rules out simple learning-rate mistuning within the tested range and strengthens the coordinate-conditioning interpretation. Specialized preconditioners and layer-specific optimization remain untested.

The same main-text table compares diagonal moments with dense $m\times m$ and $n\times n$ K-FAC factors estimated from 128--256 examples. Diagonal whitening is higher by 8.7 points on SVAMP, 0.4 on ARC-C, and 1.7 on GSM8K. Finite-sample conditioning is a plausible explanation, but this comparison does not test stronger shrinkage, structured factors, or much larger calibration sets; it only justifies the diagonal default in this regime.

\section{Resource Accounting and Reproducibility}
\label{app:repro}

\subsection{Parameter and storage formulas}
For target matrices $W_\ell\in\R^{m_\ell\times n_\ell}$, the two trainable-state counts are
\begin{align}
N_{\mathrm{core}} &= \sum_\ell r^2,\\
N_{\mathrm{factors}} &= \sum_\ell r(m_\ell+n_\ell).
\end{align}
At rank 16 in every $q/k/v/o$ projection of Qwen2.5-3B, these are 36,864 and 7,372,800, respectively, exactly a factor of 200. DoRA adds magnitude parameters for a total of 7,538,688. Adam optimizer-state savings follow the trainable-state difference, aside from framework overhead.

The complete unmerged FCCA artifact is larger than the core count: task-specific $P$ and $Q$ contain the same 7,372,800 basis elements as a pair of LoRA factors, plus the 36,864-value core. Those bases are frozen, so they do not require gradients or Adam states. If the update is merged, the separate adapter disappears at inference, but deployment then keeps a task-specific copy of the base weights. This distinction is why the paper describes FCCA as optimizer- and trainable-state efficient, not as a universally 36.9K stored adapter.

\subsection{Measured phase profile}
\begin{table}[htb]
\centering
\small
\setlength{\tabcolsep}{4.2pt}
\begin{tabular}{lrrrr}
\toprule
Method & Init./calib. & Step time & Peak GPU & Trainable \\
\midrule
FCCA & 35 s & 0.81 s & 6.92 GB & 36,864 \\
LoRA & none & 0.92 s & 7.19 GB & 7,372,800 \\
DoRA & none & 1.31 s & 7.27 GB & 7,538,688 \\
PiSSA & 41 s & 1.76 s & 7.20 GB & 7,372,800 \\
\bottomrule
\end{tabular}
\caption{Direct phase measurements on Qwen2.5-3B, SVAMP, rank 16, all $q/k/v/o$ projections, bf16, and one NVIDIA H20. The first column reports FCCA calibration and PiSSA initialization; PiSSA uses a one-time weight-SVD. ``None'' means no separate initialization or calibration phase.}
\label{tab:phase-profile}
\end{table}

\begin{table}[htb]
\centering
\scriptsize
\setlength{\tabcolsep}{2.3pt}
\begin{tabular}{lrrrr}
\toprule
Method & Init./calib. & Training & Total & Init. share \\
\midrule
FCCA & 35 s & 3.5 min & 4.1 min & 14\% \\
LoRA & none & 4.0 min & 4.0 min & n/a \\
DoRA & none & 5.7 min & 5.7 min & n/a \\
PiSSA & 41 s & 7.7 min & 8.4 min & 8\% \\
\bottomrule
\end{tabular}
\caption{Projected complete SVAMP run from measured phase times and approximately 262 optimizer steps. FCCA's first phase is calibration; PiSSA's is weight-SVD initialization. ``n/a'' means no separate first phase. Totals are projections, not independent end-to-end stopwatch measurements.}
\label{tab:projected-time}
\end{table}

Table~\ref{tab:phase-profile} reports measured phase costs, and Table~\ref{tab:projected-time} converts them into projected totals for approximately 262 optimizer steps. The one-time 35-second calibration is offset by FCCA's lower per-step time, producing a projected total close to LoRA, below DoRA, and roughly half PiSSA. Peak GPU memory differs by only 0.27--0.35 GB because the frozen model dominates; the larger saving is in gradient and optimizer state. Calibration becomes a smaller fraction of longer runs or reused bases, although host memory for cross moments can still matter at larger widths.

\subsection{Software and implementation checklist}
Measurements use a single NVIDIA H20, CUDA 12.8, PyTorch 2.7.1, Transformers 4.54, PEFT 0.16, and bf16. The reproducibility-critical settings are:
\begin{itemize}[leftmargin=*,nosep]
\item Models: Qwen2.5-1.5B/3B/7B-Instruct and Llama-3.2-1B-Instruct.
\item Seeds: 42, 43, 44; batch size 1; gradient accumulation 8.
\item Targets: every $q/k/v/o$ projection; default rank 16; zero core initialization.
\item FCCA: learning rate $5\times10^{-3}$; warmup 0.03 or 0.1; calibration 128 or 256.
\item Matched baselines: learning rate $2\times10^{-3}$ or $5\times10^{-3}$; warmup 0.1.
\item Calibration: \texttt{calib\_group\_size=999}; CPU statistics; diagonal moments by default.
\item Selection: best held-out validation configuration, followed by common evaluation on three seeds.
\item Reporting: mean $\pm$ sample standard deviation; task-level Wilcoxon on unrounded means.
\end{itemize}

\section{Use of Language Models}
A language model was used to polish the writing and assist with organizing the manuscript structure. The authors reviewed and finalized all text.

\end{document}

%% file: tables/main_q3.tex
\begin{table*}[htb]
\centering
\scriptsize
\setlength{\tabcolsep}{2.7pt}
\begin{tabular}{lcccccccc}
\toprule
Task & \textbf{FCCA} & RawGrad & FiLoRA-c & MiLoRA-c & LoRA-XS-c & VeRA-c & EVA-c & Random \\
\midrule
SVAMP & 79.7{\scriptsize$\pm$1.5} & 74.9{\scriptsize$\pm$2.0} & \textbf{80.1{\scriptsize$\pm$0.4}} & 71.2{\scriptsize$\pm$1.1} & 69.9{\scriptsize$\pm$1.8} & 72.6{\scriptsize$\pm$1.2} & 78.9{\scriptsize$\pm$1.2} & 71.2{\scriptsize$\pm$1.0} \\
GSM8K & 68.1{\scriptsize$\pm$0.8} & 66.7{\scriptsize$\pm$1.5} & 66.1{\scriptsize$\pm$2.0} & 65.2{\scriptsize$\pm$1.1} & 65.0{\scriptsize$\pm$1.6} & \textbf{69.3{\scriptsize$\pm$0.8}} & 67.8{\scriptsize$\pm$0.6} & 67.0{\scriptsize$\pm$2.7} \\
SST-2 & \textbf{95.1{\scriptsize$\pm$0.2}} & 93.6{\scriptsize$\pm$0.6} & 94.4{\scriptsize$\pm$0.2} & 93.7{\scriptsize$\pm$0.2} & 93.5{\scriptsize$\pm$0.7} & 92.3{\scriptsize$\pm$0.2} & 94.8{\scriptsize$\pm$0.3} & 92.5{\scriptsize$\pm$0.2} \\
QNLI & \textbf{88.9{\scriptsize$\pm$1.2}} & 86.9{\scriptsize$\pm$0.9} & 87.1{\scriptsize$\pm$1.1} & 87.1{\scriptsize$\pm$1.5} & 86.9{\scriptsize$\pm$1.3} & 84.8{\scriptsize$\pm$0.9} & \textbf{88.9{\scriptsize$\pm$0.7}} & 85.1{\scriptsize$\pm$0.2} \\
CoLA & 85.3{\scriptsize$\pm$1.3} & 84.1{\scriptsize$\pm$1.0} & 82.6{\scriptsize$\pm$2.0} & 85.4{\scriptsize$\pm$1.0} & 84.5{\scriptsize$\pm$0.9} & 83.9{\scriptsize$\pm$1.8} & \textbf{85.7{\scriptsize$\pm$0.1}} & 84.1{\scriptsize$\pm$0.9} \\
RTE & \textbf{89.3{\scriptsize$\pm$0.9}} & 87.5{\scriptsize$\pm$0.6} & 88.2{\scriptsize$\pm$0.7} & 88.2{\scriptsize$\pm$0.5} & \textbf{89.3{\scriptsize$\pm$1.0}} & 88.2{\scriptsize$\pm$0.5} & 88.7{\scriptsize$\pm$1.2} & 87.8{\scriptsize$\pm$0.2} \\
MRPC & \textbf{89.4{\scriptsize$\pm$1.3}} & 87.7{\scriptsize$\pm$0.7} & 87.2{\scriptsize$\pm$3.4} & 86.5{\scriptsize$\pm$0.5} & 86.4{\scriptsize$\pm$1.1} & 80.7{\scriptsize$\pm$0.3} & 81.9{\scriptsize$\pm$9.6} & 83.3{\scriptsize$\pm$0.9} \\
ARC-C & 83.1{\scriptsize$\pm$1.1} & 81.3{\scriptsize$\pm$0.9} & 82.3{\scriptsize$\pm$0.5} & 81.3{\scriptsize$\pm$0.9} & 83.0{\scriptsize$\pm$1.0} & \textbf{83.9{\scriptsize$\pm$0.7}} & 82.1{\scriptsize$\pm$1.3} & 82.5{\scriptsize$\pm$0.5} \\
OBQA & \textbf{85.9{\scriptsize$\pm$1.0}} & 75.9{\scriptsize$\pm$1.6} & 78.8{\scriptsize$\pm$0.2} & 84.7{\scriptsize$\pm$0.7} & 84.1{\scriptsize$\pm$0.8} & 85.6{\scriptsize$\pm$0.2} & 83.7{\scriptsize$\pm$1.5} & 83.7{\scriptsize$\pm$1.1} \\
HellaSwag & \textbf{75.6{\scriptsize$\pm$1.0}} & 68.2{\scriptsize$\pm$2.0} & 67.9{\scriptsize$\pm$0.3} & 74.1{\scriptsize$\pm$0.2} & 74.3{\scriptsize$\pm$1.1} & 72.4{\scriptsize$\pm$0.6} & 57.5{\scriptsize$\pm$21.2} & 71.7{\scriptsize$\pm$0.2} \\
WinoG & \textbf{72.8{\scriptsize$\pm$0.6}} & 54.0{\scriptsize$\pm$5.8} & 60.0{\scriptsize$\pm$8.0} & 67.5{\scriptsize$\pm$0.9} & 71.1{\scriptsize$\pm$1.3} & 69.5{\scriptsize$\pm$0.8} & 69.3{\scriptsize$\pm$1.7} & 68.3{\scriptsize$\pm$1.2} \\
\midrule
Mean & \textbf{83.0} & 78.3 & 79.5 & 80.4 & 80.7 & 80.3 & 79.9 & 79.7 \\
\bottomrule
\end{tabular}
\caption{Matched-budget results on Qwen2.5-3B-Instruct. Every frozen-core method trains only an $r\times r$ core with $r=16$ in all $q/k/v/o$ projections. Entries are mean$\pm$std over three seeds ($\times100$); the Mean row is the arithmetic mean of the 11 displayed task means. Bold marks the best displayed mean per row.}
\label{tab:main-q3}
\end{table*}

%% file: tables/cross_scale.tex
\begin{table*}[htb]
\centering
\small
\setlength{\tabcolsep}{4.1pt}
\begin{tabular}{lcccccccc}
\toprule
Model & \textbf{FCCA} & RawGrad & FiLoRA-c & MiLoRA-c & LoRA-XS-c & VeRA-c & EVA-c & Random \\
\midrule
Qwen2.5-1.5B & \textbf{77.2} & 75.3 & 75.2 & 74.9 & 74.8 & 74.8 & \underline{76.7} & 73.8 \\
Qwen2.5-3B & \textbf{83.0} & 78.3 & 79.5 & 80.4 & \underline{80.7} & 80.3 & 79.9 & 79.7 \\
Qwen2.5-7B & \textbf{87.5} & 86.0 & 85.5 & 84.3 & 84.9 & 85.7 & \underline{86.7} & 82.9 \\
Llama-3.2-1B & \underline{68.1} & 66.3 & 67.5 & 64.7 & 66.6 & 66.4 & \textbf{68.2} & 63.1 \\
\bottomrule
\end{tabular}
\caption{Cross-scale 11-task means under the same protocol. Values are recomputed from the displayed per-task means; bold/underline denote first/second within each model.}
\label{tab:cross-scale}
\end{table*}

%% file: tables/full_q15.tex
\begin{table*}[htb]
\centering
\scriptsize
\setlength{\tabcolsep}{2.7pt}
\begin{tabular}{lcccccccc}
\toprule
Task & \textbf{FCCA} & RawGrad & FiLoRA-c & MiLoRA-c & LoRA-XS-c & VeRA-c & EVA-c & Random \\
\midrule
SVAMP & 68.4{\scriptsize$\pm$1.0} & 69.1{\scriptsize$\pm$2.0} & 67.2{\scriptsize$\pm$0.6} & 62.7{\scriptsize$\pm$1.0} & 59.1{\scriptsize$\pm$1.1} & 64.3{\scriptsize$\pm$3.4} & \textbf{69.2{\scriptsize$\pm$0.8}} & 59.4{\scriptsize$\pm$1.5} \\
GSM8K & 47.3{\scriptsize$\pm$0.7} & 45.7{\scriptsize$\pm$1.2} & \textbf{49.1{\scriptsize$\pm$1.8}} & 46.8{\scriptsize$\pm$1.1} & 46.4{\scriptsize$\pm$0.8} & 48.1{\scriptsize$\pm$0.9} & 48.4{\scriptsize$\pm$1.5} & 46.4{\scriptsize$\pm$0.9} \\
SST-2 & \textbf{95.8{\scriptsize$\pm$0.3}} & 93.7{\scriptsize$\pm$0.8} & 93.7{\scriptsize$\pm$0.7} & 94.8{\scriptsize$\pm$0.2} & 94.9{\scriptsize$\pm$0.3} & 94.4{\scriptsize$\pm$0.2} & 95.0{\scriptsize$\pm$0.7} & 94.4{\scriptsize$\pm$0.4} \\
QNLI & \textbf{87.5{\scriptsize$\pm$0.5}} & 87.1{\scriptsize$\pm$0.4} & 85.7{\scriptsize$\pm$0.7} & 85.8{\scriptsize$\pm$0.9} & 85.7{\scriptsize$\pm$0.6} & 84.9{\scriptsize$\pm$0.5} & 86.4{\scriptsize$\pm$0.9} & 84.9{\scriptsize$\pm$0.7} \\
CoLA & 82.2{\scriptsize$\pm$1.1} & \textbf{82.6{\scriptsize$\pm$1.3}} & 80.9{\scriptsize$\pm$2.6} & 80.2{\scriptsize$\pm$0.7} & 80.1{\scriptsize$\pm$1.2} & 75.7{\scriptsize$\pm$1.5} & 82.5{\scriptsize$\pm$2.3} & 74.5{\scriptsize$\pm$1.1} \\
RTE & \textbf{85.9{\scriptsize$\pm$1.6}} & 84.1{\scriptsize$\pm$1.2} & 81.3{\scriptsize$\pm$0.7} & 82.4{\scriptsize$\pm$0.7} & 84.0{\scriptsize$\pm$1.1} & 82.2{\scriptsize$\pm$0.2} & 85.0{\scriptsize$\pm$0.9} & 82.6{\scriptsize$\pm$0.9} \\
MRPC & \textbf{87.5{\scriptsize$\pm$0.3}} & 85.6{\scriptsize$\pm$0.6} & 85.0{\scriptsize$\pm$0.9} & 84.2{\scriptsize$\pm$0.9} & 85.3{\scriptsize$\pm$0.9} & 84.2{\scriptsize$\pm$0.3} & 87.3{\scriptsize$\pm$1.1} & 82.8{\scriptsize$\pm$0.4} \\
ARC-C & \textbf{77.7{\scriptsize$\pm$1.6}} & 74.4{\scriptsize$\pm$1.2} & 76.5{\scriptsize$\pm$1.6} & 75.8{\scriptsize$\pm$1.6} & 76.6{\scriptsize$\pm$1.3} & 77.4{\scriptsize$\pm$0.7} & 76.0{\scriptsize$\pm$1.1} & 76.8{\scriptsize$\pm$0.9} \\
OBQA & \textbf{82.1{\scriptsize$\pm$0.4}} & 78.9{\scriptsize$\pm$1.3} & 79.8{\scriptsize$\pm$2.0} & 81.1{\scriptsize$\pm$0.3} & 80.3{\scriptsize$\pm$0.6} & 81.9{\scriptsize$\pm$0.5} & 82.0{\scriptsize$\pm$0.3} & 79.9{\scriptsize$\pm$0.2} \\
HellaSwag & \textbf{68.1{\scriptsize$\pm$0.5}} & 65.1{\scriptsize$\pm$0.3} & 64.9{\scriptsize$\pm$1.5} & 65.9{\scriptsize$\pm$0.6} & 65.8{\scriptsize$\pm$0.8} & 65.7{\scriptsize$\pm$1.2} & 67.1{\scriptsize$\pm$0.5} & 67.1{\scriptsize$\pm$0.2} \\
WinoG & \textbf{66.2{\scriptsize$\pm$0.3}} & 61.5{\scriptsize$\pm$1.5} & 63.3{\scriptsize$\pm$1.2} & 63.8{\scriptsize$\pm$0.7} & 64.9{\scriptsize$\pm$1.1} & 64.1{\scriptsize$\pm$1.4} & 64.6{\scriptsize$\pm$1.8} & 62.5{\scriptsize$\pm$0.8} \\
\midrule
Mean & \textbf{77.2} & 75.3 & 75.2 & 74.9 & 74.8 & 74.8 & 76.7 & 73.8 \\
\bottomrule
\end{tabular}
\caption{Full matched-budget results on Qwen2.5-1.5B-Instruct. Mean$\pm$std over three seeds ($\times100$); the Mean row is recomputed from the 11 displayed task means.}
\label{tab:full-q15}
\end{table*}

%% file: tables/full_q7.tex
\begin{table*}[htb]
\centering
\scriptsize
\setlength{\tabcolsep}{2.7pt}
\begin{tabular}{lcccccccc}
\toprule
Task & \textbf{FCCA} & RawGrad & FiLoRA-c & MiLoRA-c & LoRA-XS-c & VeRA-c & EVA-c & Random \\
\midrule
SVAMP & \textbf{84.4{\scriptsize$\pm$0.4}} & 83.8{\scriptsize$\pm$0.3} & 81.6{\scriptsize$\pm$0.6} & 76.6{\scriptsize$\pm$0.2} & 78.7{\scriptsize$\pm$1.7} & 80.1{\scriptsize$\pm$2.0} & 81.7{\scriptsize$\pm$0.5} & 75.7{\scriptsize$\pm$1.1} \\
GSM8K & 78.1{\scriptsize$\pm$0.6} & 77.2{\scriptsize$\pm$0.4} & 75.9{\scriptsize$\pm$0.6} & 73.0{\scriptsize$\pm$1.1} & 72.2{\scriptsize$\pm$1.7} & \textbf{78.2{\scriptsize$\pm$1.0}} & 77.3{\scriptsize$\pm$0.5} & 72.0{\scriptsize$\pm$0.7} \\
SST-2 & \textbf{96.4{\scriptsize$\pm$0.0}} & 95.9{\scriptsize$\pm$0.2} & 94.9{\scriptsize$\pm$0.8} & 95.1{\scriptsize$\pm$0.2} & 95.1{\scriptsize$\pm$0.2} & 94.9{\scriptsize$\pm$0.1} & 96.2{\scriptsize$\pm$0.2} & 94.8{\scriptsize$\pm$0.3} \\
QNLI & 89.4{\scriptsize$\pm$0.9} & 87.6{\scriptsize$\pm$0.2} & 89.2{\scriptsize$\pm$0.6} & 88.1{\scriptsize$\pm$0.7} & 88.4{\scriptsize$\pm$0.3} & 88.3{\scriptsize$\pm$0.7} & \textbf{90.2{\scriptsize$\pm$0.7}} & 87.3{\scriptsize$\pm$0.6} \\
CoLA & 86.8{\scriptsize$\pm$0.3} & 86.8{\scriptsize$\pm$1.6} & 85.9{\scriptsize$\pm$1.0} & 85.9{\scriptsize$\pm$0.2} & 86.1{\scriptsize$\pm$0.2} & 85.3{\scriptsize$\pm$0.4} & \textbf{87.4{\scriptsize$\pm$1.0}} & 83.3{\scriptsize$\pm$1.0} \\
RTE & \textbf{91.5{\scriptsize$\pm$0.5}} & 90.1{\scriptsize$\pm$0.6} & 90.1{\scriptsize$\pm$0.7} & 90.3{\scriptsize$\pm$0.9} & 90.6{\scriptsize$\pm$0.8} & 91.0{\scriptsize$\pm$0.8} & 90.6{\scriptsize$\pm$1.3} & 89.2{\scriptsize$\pm$0.3} \\
MRPC & \textbf{88.5{\scriptsize$\pm$0.4}} & 86.6{\scriptsize$\pm$1.5} & 87.7{\scriptsize$\pm$0.3} & 84.7{\scriptsize$\pm$0.6} & 87.4{\scriptsize$\pm$0.5} & 86.4{\scriptsize$\pm$0.8} & 88.1{\scriptsize$\pm$0.4} & 83.3{\scriptsize$\pm$0.9} \\
ARC-C & \textbf{91.2{\scriptsize$\pm$0.4}} & 89.7{\scriptsize$\pm$0.5} & 88.5{\scriptsize$\pm$0.7} & \textbf{91.2{\scriptsize$\pm$0.3}} & 88.9{\scriptsize$\pm$0.2} & 90.6{\scriptsize$\pm$0.2} & 91.0{\scriptsize$\pm$0.3} & 89.7{\scriptsize$\pm$0.3} \\
OBQA & \textbf{92.5{\scriptsize$\pm$0.7}} & 89.8{\scriptsize$\pm$0.7} & 90.1{\scriptsize$\pm$0.5} & 89.3{\scriptsize$\pm$0.1} & 89.5{\scriptsize$\pm$0.8} & 90.3{\scriptsize$\pm$0.4} & 90.3{\scriptsize$\pm$0.9} & 87.4{\scriptsize$\pm$0.3} \\
HellaSwag & \textbf{82.7{\scriptsize$\pm$0.5}} & 80.1{\scriptsize$\pm$0.1} & 80.1{\scriptsize$\pm$1.1} & 79.1{\scriptsize$\pm$0.9} & 81.2{\scriptsize$\pm$1.2} & 81.7{\scriptsize$\pm$1.0} & 81.8{\scriptsize$\pm$0.3} & 79.4{\scriptsize$\pm$0.6} \\
WinoG & \textbf{80.9{\scriptsize$\pm$0.8}} & 78.1{\scriptsize$\pm$1.2} & 76.3{\scriptsize$\pm$1.1} & 73.9{\scriptsize$\pm$0.8} & 75.8{\scriptsize$\pm$0.4} & 75.5{\scriptsize$\pm$0.3} & 79.2{\scriptsize$\pm$0.8} & 69.3{\scriptsize$\pm$1.1} \\
\midrule
Mean & \textbf{87.5} & 86.0 & 85.5 & 84.3 & 84.9 & 85.7 & 86.7 & 82.9 \\
\bottomrule
\end{tabular}
\caption{Full matched-budget results on Qwen2.5-7B-Instruct. Mean$\pm$std over three seeds ($\times100$); the Mean row is recomputed from the 11 displayed task means.}
\label{tab:full-q7}
\end{table*}

%% file: tables/full_llama1.tex
\begin{table*}[htb]
\centering
\scriptsize
\setlength{\tabcolsep}{2.7pt}
\begin{tabular}{lcccccccc}
\toprule
Task & \textbf{FCCA} & RawGrad & FiLoRA-c & MiLoRA-c & LoRA-XS-c & VeRA-c & EVA-c & Random \\
\midrule
SVAMP & \textbf{64.9{\scriptsize$\pm$1.5}} & 61.7{\scriptsize$\pm$0.3} & \textbf{64.9{\scriptsize$\pm$1.0}} & 55.9{\scriptsize$\pm$0.3} & 58.3{\scriptsize$\pm$1.2} & 63.2{\scriptsize$\pm$0.2} & 62.2{\scriptsize$\pm$0.7} & 55.3{\scriptsize$\pm$2.2} \\
GSM8K & 38.6{\scriptsize$\pm$1.7} & \textbf{38.8{\scriptsize$\pm$0.3}} & 36.0{\scriptsize$\pm$0.9} & 37.0{\scriptsize$\pm$0.5} & 35.1{\scriptsize$\pm$1.8} & 37.7{\scriptsize$\pm$2.2} & 36.1{\scriptsize$\pm$0.7} & 34.7{\scriptsize$\pm$1.4} \\
SST-2 & 87.5{\scriptsize$\pm$4.4} & 86.5{\scriptsize$\pm$7.7} & 90.6{\scriptsize$\pm$2.4} & 91.0{\scriptsize$\pm$1.2} & 92.3{\scriptsize$\pm$0.3} & 92.5{\scriptsize$\pm$0.5} & \textbf{93.3{\scriptsize$\pm$1.4}} & 91.0{\scriptsize$\pm$0.7} \\
QNLI & \textbf{85.1{\scriptsize$\pm$1.2}} & 81.3{\scriptsize$\pm$2.0} & 83.8{\scriptsize$\pm$1.4} & 77.2{\scriptsize$\pm$1.4} & 81.4{\scriptsize$\pm$2.5} & 77.1{\scriptsize$\pm$1.9} & 84.3{\scriptsize$\pm$1.6} & 75.1{\scriptsize$\pm$0.8} \\
CoLA & 73.9{\scriptsize$\pm$3.7} & \textbf{74.5{\scriptsize$\pm$3.2}} & 72.1{\scriptsize$\pm$1.5} & 70.0{\scriptsize$\pm$0.0} & 70.3{\scriptsize$\pm$0.4} & 70.0{\scriptsize$\pm$0.0} & 73.7{\scriptsize$\pm$1.9} & 70.0{\scriptsize$\pm$0.0} \\
RTE & 78.7{\scriptsize$\pm$1.4} & 75.1{\scriptsize$\pm$1.5} & \textbf{79.4{\scriptsize$\pm$0.0}} & 76.3{\scriptsize$\pm$1.8} & 76.7{\scriptsize$\pm$1.4} & 75.0{\scriptsize$\pm$1.2} & 78.2{\scriptsize$\pm$0.9} & 70.5{\scriptsize$\pm$0.9} \\
MRPC & 80.4{\scriptsize$\pm$0.7} & 79.5{\scriptsize$\pm$3.6} & 80.3{\scriptsize$\pm$0.6} & 74.6{\scriptsize$\pm$0.8} & 79.1{\scriptsize$\pm$1.1} & 75.3{\scriptsize$\pm$1.3} & \textbf{80.7{\scriptsize$\pm$2.8}} & 70.8{\scriptsize$\pm$1.8} \\
ARC-C & 57.4{\scriptsize$\pm$0.6} & 58.3{\scriptsize$\pm$0.7} & 57.4{\scriptsize$\pm$0.7} & 57.7{\scriptsize$\pm$0.9} & 60.7{\scriptsize$\pm$1.2} & \textbf{61.5{\scriptsize$\pm$0.2}} & 60.9{\scriptsize$\pm$1.3} & 58.1{\scriptsize$\pm$0.9} \\
OBQA & 68.3{\scriptsize$\pm$1.0} & 67.0{\scriptsize$\pm$1.0} & 67.9{\scriptsize$\pm$0.4} & 66.1{\scriptsize$\pm$0.5} & 67.8{\scriptsize$\pm$0.9} & \textbf{69.6{\scriptsize$\pm$0.3}} & 69.5{\scriptsize$\pm$0.9} & 65.3{\scriptsize$\pm$1.0} \\
HellaSwag & 57.0{\scriptsize$\pm$1.0} & 54.5{\scriptsize$\pm$2.6} & 56.5{\scriptsize$\pm$0.7} & 53.9{\scriptsize$\pm$0.6} & 55.5{\scriptsize$\pm$1.7} & 57.1{\scriptsize$\pm$1.5} & \textbf{59.1{\scriptsize$\pm$1.0}} & 53.0{\scriptsize$\pm$1.8} \\
WinoG & \textbf{56.9{\scriptsize$\pm$2.0}} & 52.6{\scriptsize$\pm$3.1} & 53.3{\scriptsize$\pm$4.0} & 52.0{\scriptsize$\pm$1.6} & 55.2{\scriptsize$\pm$1.7} & 51.2{\scriptsize$\pm$1.3} & 52.1{\scriptsize$\pm$2.9} & 50.4{\scriptsize$\pm$0.0} \\
\midrule
Mean & 68.1 & 66.3 & 67.5 & 64.7 & 66.6 & 66.4 & \textbf{68.2} & 63.1 \\
\bottomrule
\end{tabular}
\caption{Full matched-budget results on Llama-3.2-1B-Instruct. Mean$\pm$std over three seeds ($\times100$); the Mean row is recomputed from the 11 displayed task means.}
\label{tab:full-llama1}
\end{table*}

%% file: tables/trainable_basis_full.tex
\begin{table}[htb]
\centering
\scriptsize
\setlength{\tabcolsep}{1.3pt}
\begin{tabular}{lcccc}
\toprule
Task & FCCA (36.9K) & LoRA (7.37M) & DoRA (7.54M) & PiSSA (7.37M) \\
\midrule
SVAMP & \textbf{79.7{\scriptsize$\pm$1.5}} & 78.8{\scriptsize$\pm$1.6} & 79.2{\scriptsize$\pm$0.3} & 77.3{\scriptsize$\pm$2.6} \\
GSM8K & \textbf{68.1{\scriptsize$\pm$0.8}} & 67.7{\scriptsize$\pm$1.9} & 67.2{\scriptsize$\pm$0.7} & 51.4{\scriptsize$\pm$1.7} \\
SST-2 & \textbf{95.1{\scriptsize$\pm$0.2}} & 94.5{\scriptsize$\pm$0.5} & 94.9{\scriptsize$\pm$0.4} & 94.4{\scriptsize$\pm$0.7} \\
QNLI & \textbf{88.9{\scriptsize$\pm$1.2}} & 87.9{\scriptsize$\pm$0.6} & 87.4{\scriptsize$\pm$0.6} & 87.3{\scriptsize$\pm$1.5} \\
CoLA & 85.3{\scriptsize$\pm$1.3} & \textbf{85.9{\scriptsize$\pm$0.7}} & \textbf{85.9{\scriptsize$\pm$0.7}} & 85.0{\scriptsize$\pm$1.0} \\
RTE & 89.3{\scriptsize$\pm$0.9} & \textbf{89.7{\scriptsize$\pm$0.2}} & 89.4{\scriptsize$\pm$0.7} & 88.6{\scriptsize$\pm$2.1} \\
MRPC & 89.4{\scriptsize$\pm$1.3} & \textbf{89.5{\scriptsize$\pm$0.4}} & 89.4{\scriptsize$\pm$0.3} & 89.1{\scriptsize$\pm$0.5} \\
ARC-C & 83.1{\scriptsize$\pm$1.1} & 84.3{\scriptsize$\pm$0.3} & \textbf{84.7{\scriptsize$\pm$0.5}} & 82.5{\scriptsize$\pm$0.8} \\
OBQA & 85.9{\scriptsize$\pm$1.0} & 88.2{\scriptsize$\pm$0.3} & \textbf{88.3{\scriptsize$\pm$0.7}} & 84.6{\scriptsize$\pm$1.2} \\
HellaSwag & 75.6{\scriptsize$\pm$1.0} & \textbf{76.9{\scriptsize$\pm$1.1}} & 76.6{\scriptsize$\pm$1.3} & 73.0{\scriptsize$\pm$0.2} \\
WinoG & 72.8{\scriptsize$\pm$0.6} & \textbf{73.3{\scriptsize$\pm$1.3}} & 72.7{\scriptsize$\pm$0.9} & 71.4{\scriptsize$\pm$0.7} \\
\midrule
Mean & 83.0 & \textbf{83.3} & 83.2 & 80.4 \\
\bottomrule
\end{tabular}
\caption{Full Qwen2.5-3B comparison with trainable-basis rank-16 references. Mean$\pm$std over three seeds ($\times100$).}
\label{tab:trainable-basis-full}
\end{table}